\documentclass[sigplan]{acmart}

\usepackage[utf8]{inputenc}
\usepackage[T1]{fontenc}
\usepackage{amsmath}
 
\usepackage{amssymb}
\usepackage{amsfonts}
\usepackage{algorithm,algpseudocode}
\usepackage{graphicx}
\usepackage{booktabs}
\usepackage{hyperref}
\usepackage{xcolor}
\usepackage{tikz}
\usepackage{caption}
\usepackage{subcaption}
\usepackage{listings}
\usepackage{multirow}
\usepackage{url}

\hypersetup{
    colorlinks=true,
    linkcolor=blue,
    filecolor=magenta,      
    urlcolor=cyan,
    citecolor=blue
}

\usepackage{tabularx} 
\usepackage{graphicx}
\usepackage[T1]{fontenc}
\usepackage{enumitem}
\usepackage{soul}
\usepackage{url}
\usepackage{subcaption}
\usepackage{array} 
\usepackage{booktabs} 

\usepackage{xspace}
\usepackage{xcolor}
\usepackage{wrapfig} 
\usepackage{caption} 
\usepackage{booktabs} 
\usepackage{tikz}

\usepackage{listings}
\usepackage{xcolor}

\usetikzlibrary{positioning,arrows.meta,calc}

\newcommand{\para}[1]{\medskip\noindent\textbf{#1}}

\newenvironment{parafont}{\fontfamily{ptm}\selectfont}{}

\AtBeginDocument{%
  }

\newcommand{\tightcaption}[1]{\vspace{-7pt}\caption{{\bf \small #1}}
\vspace{-8pt}
}

\setcopyright{none}
\renewcommand\footnotetextcopyrightpermission[1]{} 
\begin{document}

\title{\sys: Speculative Execution for Fast and Efficient Web Agents}

\author{Mike Wong}
\affiliation{%
  \institution{Princeton University}
    \city{Princeton}
  \country{USA}
}
\author{Kevin Hsieh}
\affiliation{%
  \institution{Microsoft Research}
    \city{Seattle}
  \country{USA}
}
\author{Suman Nath}
\affiliation{%
  \institution{Microsoft Research}
    \city{Seattle}
  \country{USA}
}
\author{Ravi Netravali}
\affiliation{%
  \institution{Princeton University}
   \city{Princeton}
  \country{USA}
}

\renewcommand{\shortauthors}{M. Wong et al.}
\newcommand{\sys}{{Skim}\xspace} 


\addtolength{\textheight}{-0.25in}


\begin{abstract}


Skim is a speculative execution framework for web agents that exploits the predictable structure of purpose-built websites. Today's web-agent expense is not intrinsic to the tasks but a property of how agents are composed: frontier-model inference, browser rendering, and ReAct-style planning are applied to every step of every task regardless of complexity. Skim's key observation is that websites enforce stable URL patterns, answer formats, and task-to-trajectory mappings across queries of the same type, so most queries can bypass these heavyweight components entirely. An offline profiler captures these patterns once per site. At runtime, Skim matches each query to a template, synthesizes the destination URL, and extracts the answer with a small model. A lightweight verifier gates each fast-path output against the query and schema; rare misspeculations cascade to the full agent, warm-started by the fast path's final URL to preserve upstream trajectory progress. Across standard web-agent benchmarks paired with three backboneagents (WebVoyager, AgentOccam, BrowserUse), Skim reduces median per-task cost by 1.9x and latency by 33.4\% with no accuracy loss.

\end{abstract}

\keywords{}

\received{20 February 2007}
\received[revised]{12 March 2009}
\received[accepted]{5 June 2009}

\maketitle

\newcommand{\todo}[1]{\textcolor{red}{[TODO: #1]}}
\newcommand{\note}[1]{\textcolor{blue}{\textit{Note: #1}}}
\newcommand{\figplaceholder}[2]{%
  \begin{center}
  \fbox{\parbox{0.85\linewidth}{\centering\vspace{1em}\textbf{Figure \ref*{#1}: #2}\\[0.5em]\textcolor{gray}{[placeholder]}\vspace{1em}}}
  \end{center}
}

\section{Introduction}

\begin{sloppypar}
Web agents, or LLM-based systems that complete tasks by directly browsing and interacting with live websites, are increasingly deployed in retrieval-augmented assistants, deep-research tools, and enterprise automation pipelines. Agents such as WebVoyager~\cite{webvoyager}, AgentOccam~\cite{agentoccam}, and BrowserUse~\cite{browseruse} now underpin a fast-growing class of products that resolve queries no static corpus can answer: navigating to authenticated pages, completing multi-step lookups across catalogs, and extracting structured information from sites the user does not need to visit themselves. This capability far exceeds what stateless retrieval pipelines (e.g., Perplexity, ChatGPT browsing) achieve by querying search indices and synthesizing from snippets -- they cannot maintain authenticated state, complete multi-step trajectories, or reach content beyond search coverage -- but it comes at substantial overhead: per-task latencies run 30-120 secs and API costs run \$0.20-0.50, 1-2 orders of magnitude more than stateless retrieval. 
\end{sloppypar}

The high overheads of web agents trace to three off-the-shelf components, each designed for problems broader than the web tasks they typically execute. Frontier LLMs are trained for long-context multi-hop reasoning; browsers are engineered for arbitrary and secure human interaction with any web page; ReAct is a general decision-making framework for agents in unconstrained environments with sparse feedback. Each component is highly capable, but fails to exploit the structural regularities that the web exhibits. The first such property is \emph{step heterogeneity}. Most steps in a typical web agent trajectory are mechanical (fetch a known URL, locate a labeled field on a page, click a stable navigation element) and do not require frontier reasoning or full browser execution. The second is \emph{task-driven structure}. Sites are purpose-built around the queries users bring to them, so URL templates, page layouts, and answer formats are stable across queries of the same type. For instance, the navigation pattern that resolves one arXiv lookup resolves the next, with only query-specific identifiers changing. Together, these properties imply that many steps can run on cheaper components, and many can be skipped entirely by following trajectories the site has already revealed.

Our analysis of popular web agent tasks confirms that the opportunity is substantial (\S\ref{sec:opp}). Hand-engineered programs that exploit website structure by using trajectory knowledge to fetch via direct URLs, prune to answer-bearing content, and extract with cheaper models can run 66.7-94.9\% faster and 17.7-100.7$\times$ cheaper than the off-the-shelf ReAct agent without any accuracy loss. However, automating this strategy in practice is difficult; naively substituting these cheaper techniques inside an off-the-shelf ReAct loop drops average task success by 60\%. The issue is that cheap models struggle to choose the right action from the hundreds of interactive elements on a typical page, and direct-URL fetching requires knowing in advance precisely which page will contain the answer for a given query (\S\ref{sec:chal}). Worse, errors compound across a trajectory, so a single wrong step can invalidate everything downstream. The core question is thus how to identify, when a query arrives, which shortcuts preserve accuracy for \emph{that} query -- a question made difficult by the heterogeneity of tasks, the variation across pages in a site, and the per-step diversity in resource needs.


To address these challenges, we present \textbf{\sys}, a drop-in acceleration framework for existing web agents. The key insight underlying \sys{} is that web sites are \emph{predictable enough to precompute} navigation paths offline, and \emph{structured enough to check} answer correctness online. Sites change rarely relative to query rates, so trajectory patterns characterized once per task type remain valid across many future queries. Further, since sites impose stable formats on task result pages (e.g., search results), a fast path's output can be easily validated against both the query and the expected schema, without invoking the expensive agent. Building on this, \sys{} profiles each reachable site offline, recording for each task type both the shortcuts that apply (URL patterns reaching its result pages) and what correct outputs look like (the answer schemas of those pages). At runtime, when a query arrives, \sys{} consults the profile and, if covered, speculatively attempts a fast path via direct URL fetch and cheap model processing, accepting the result only when a small judge model deems it consistent with the query and the recorded schema. Upon failure, \sys{} falls back to the full ReAct agent, warm-started at the URL the fast path reached -- since trajectories share predictable prefixes, even failed shortcuts reduce the work that the agent must redo.


We evaluate \sys on two representative benchmarks (WebVoyager, WebShop) across three popular ReAct-based web agents (BrowserUse, AgentOccam, WebVoyager). We consider two deployment modes that exploit the same underlying mechanism. In \emph{accelerate} mode, \sys lowers median per-task cost by 1.9$\times$ and latency by 33.4\% while preserving accuracy. In \emph{aggregate} mode, \sys reinvests the saved compute budget into multiple trials that explore different trajectories with verifier-selected outputs. Within the single-trial baseline's cost budget, this lifts end-to-end accuracy by up to 16.7 percentage points (4.2 pp with majority vote).

\begin{figure}[t]
    \centering
    \includegraphics[width=0.8\columnwidth]{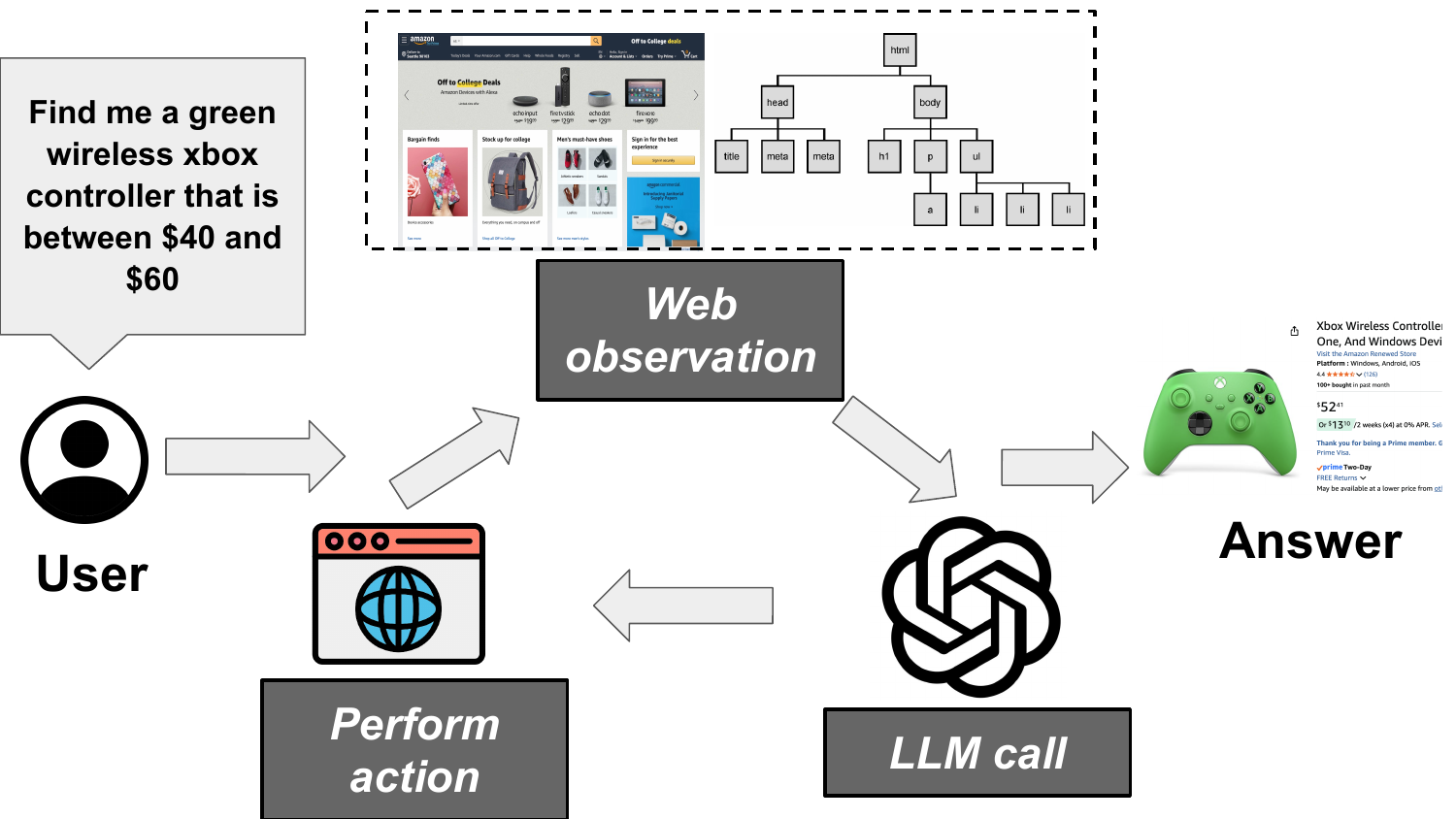}
    \tightcaption{Workflow of a representative ReAct-based web agent.
    The agent loops through render, observe, reason, and act steps
    until task completion.}
    \label{fig:react-workflow}
\end{figure}

\section{Background and Motivation}
\label{sec:background}

Web agents incur substantial overheads not intrinsic to their tasks. We examine where these overheads come from, identify the structural properties that make them avoidable, and characterize the challenges of automating their removal.

\begin{figure}[t]
    \centering
    \includegraphics[width=0.8\columnwidth]{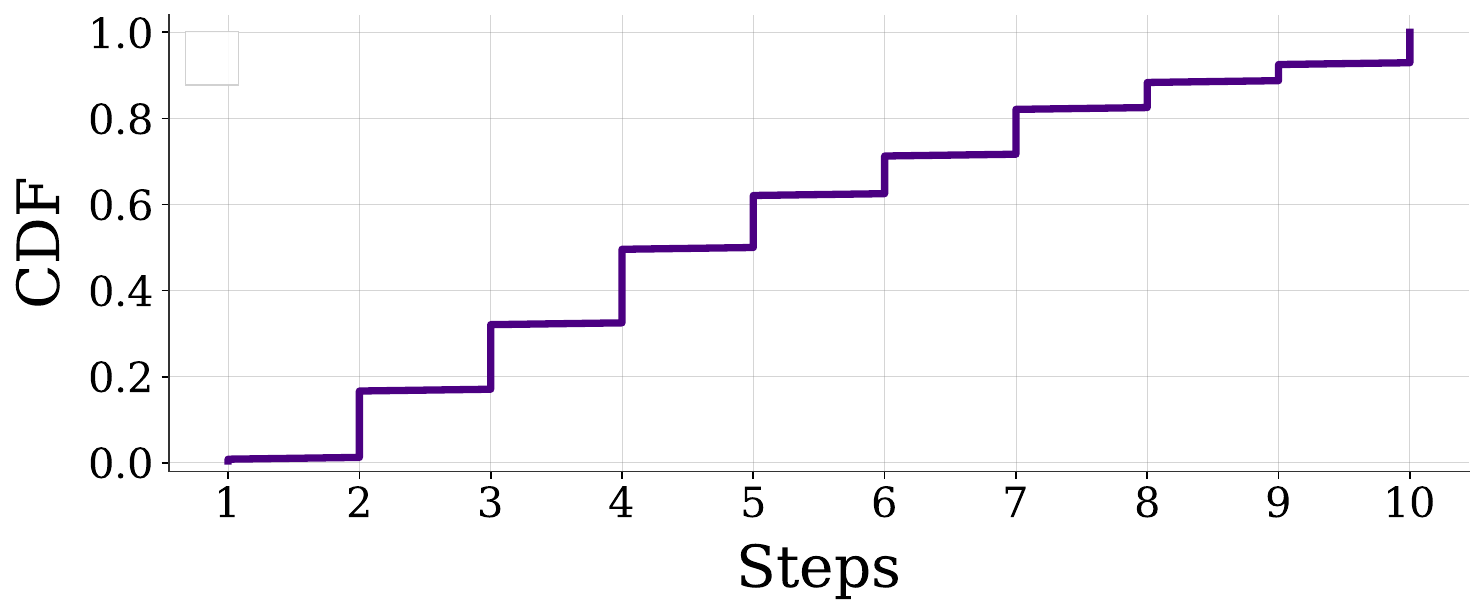}
    \tightcaption{Distribution of number of ReAct steps needed for agents to converge to an answer.}
    \label{fig:steps}
\end{figure}

\begin{figure}[t]
    \centering
    \includegraphics[width=0.8\columnwidth]{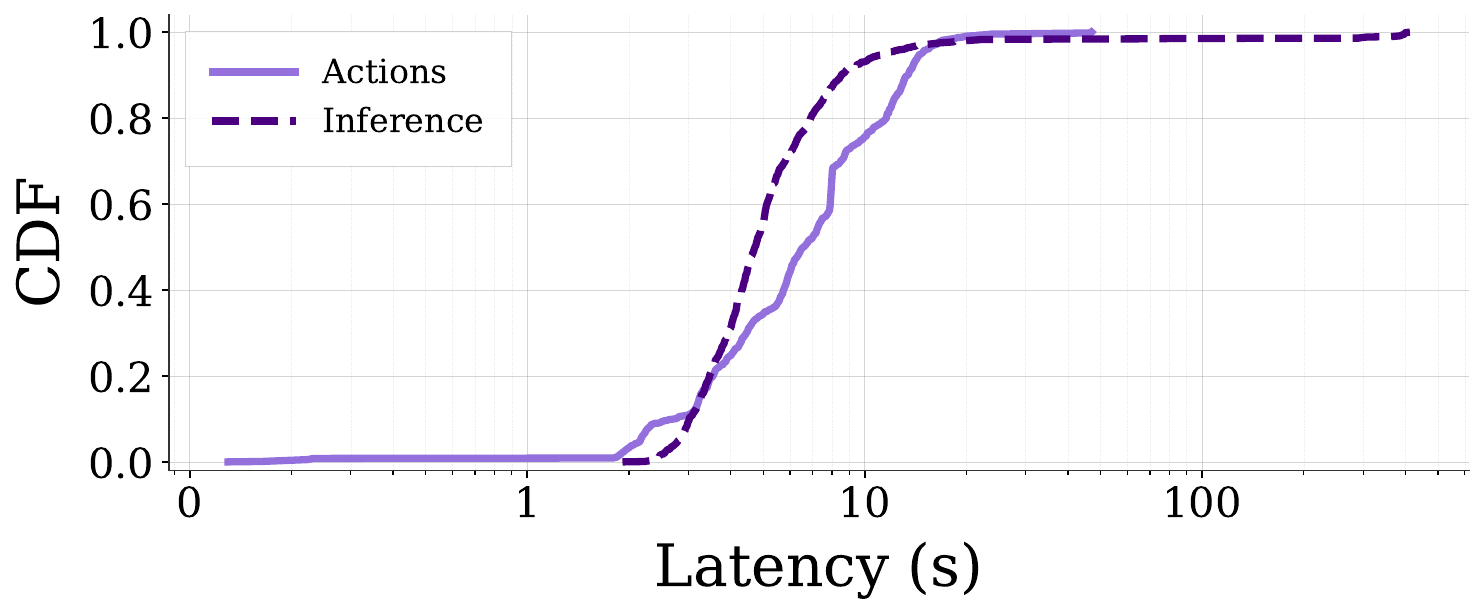}
    \tightcaption{Per-step breakdown of latencies. Actions involve typing, clicking, googling, and scrolling while inference involves running the frontier model on the web input}
    \label{fig:latency-breakdown}
\end{figure}

\subsection{Overview of web agents}
\label{sec:bg}

Canonical web agents such as WebVoyager~\cite{webvoyager}, AgentOccam~\cite{agentoccam}, BrowserUse~\cite{browseruse}, and SeeAct~\cite{seeact} operate by coupling a frontier LLM with a headless browser through a ReAct execution loop~\cite{react} (Figure~\ref{fig:react-workflow}). At each step, the agent loads the current page in the browser, observes its rendered representation as either a serialized DOM tree or a screenshot, reasons about the next action to take, and executes that action against the page through standard browser primitives such as clicks, typed input, scrolling, and navigation. Every step requires a full LLM call that processes the current page representation alongside the running interaction history. The loop repeats until the agent decides the task is complete, deemed unachievable, or until a step budget is exhausted.

Web agents are designed for tasks that require navigating human-facing sites to retrieve targeted information. Example tasks include looking up product prices and availability across e-commerce catalogs, retrieving paper abstracts and citation counts from academic repositories, comparing course offerings across educational platforms, summarizing freshly-published news articles, and gathering reservation details from booking sites. They differ from tool-using agents that call structured APIs or MCP servers designed for programmatic access (e.g., coding agents acting on local files and developer tools~\cite{swt-bench, execution-agent}), and from stateless retrieval pipelines that synthesize answers from search-engine results without visiting the underlying pages~\cite{caesar}. The defining feature of a web agent is that it operates against the \emph{live web}, and must reason about page content and interaction patterns that vary across sites and across queries.

\subsection{Opportunities for specialization}
\label{sec:opp}

We start by profiling overheads across 151 representative tasks from the WebVoyager benchmark using three web agents -- BrowserUse, AgentOccam, WebVoyager -- and the experimental methodology outlined in \S\ref{sev:eval}.  Overall, we find that per-task latencies span 30-120 seconds, translating to per-task API costs of \$0.20-0.50 for model inference. Digging deeper, this overhead decomposes into two factors that each contribute substantially: the number of steps a trajectory takes, and the overhead of each step. As Figure~\ref{fig:steps} shows, the median task takes 4 steps with 80\% of tasks requiring at least 7 steps. Figure~\ref{fig:latency-breakdown} breaks down each step into browser actions and LLM inference delays, showing that both contribute meaningfully (median delays of 4.7 secs for LLM inference and 6.6 secs for browser action time, respectively). Since action decisions are conditioned on current page content and the trajectory thus far, per-step inference costs grow as history builds.

We argue that these overheads are not intrinsic to the underlying web agent tasks. Instead, they stem from applying heavy, general-purpose components -- frontier-model inference, full browser rendering, and the iterative ReAct loop, each designed for problems much broader than typical web agent tasks -- \emph{uniformly} across all steps regardless of need. We next describe three intrinsic properties of the web that make this uniformity unnecessary, and that together open the door to substantial specialization for acceleration.

\begin{figure}[t]
    \centering
    \includegraphics[width=0.8\columnwidth]{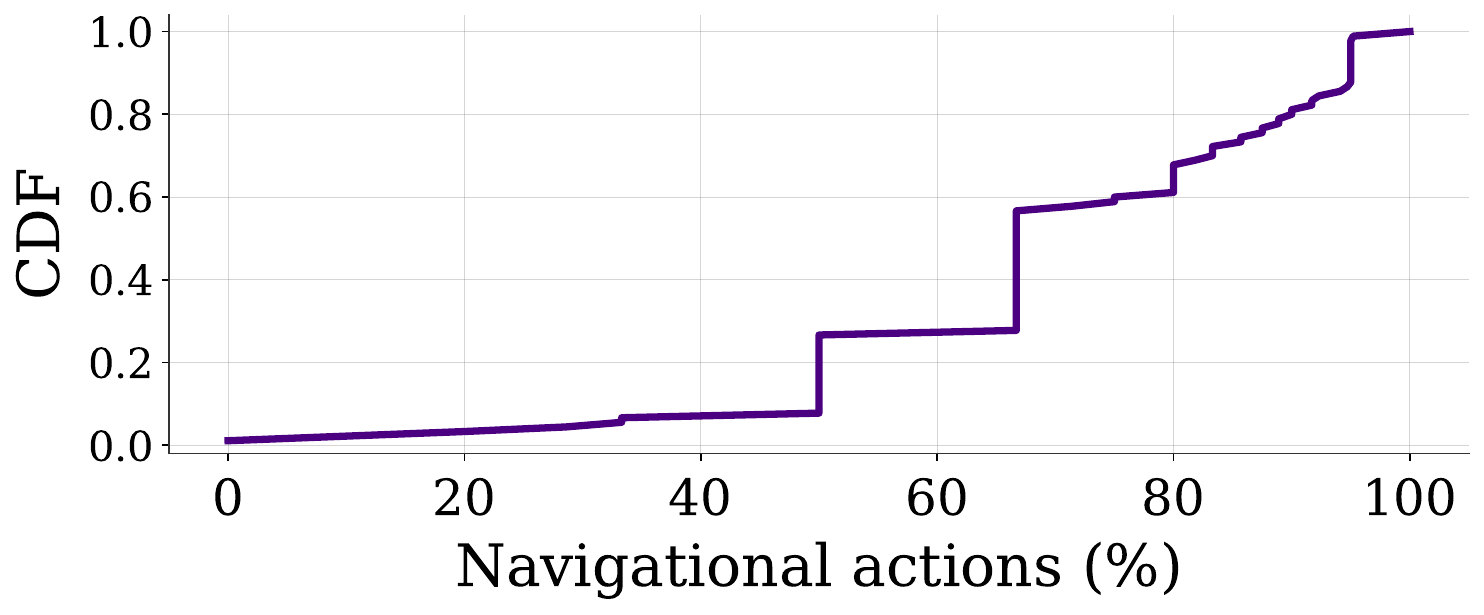}
    \tightcaption{Distribution of the percentage of steps per task that are navigational, i.e., moving the agent between pages rather than directly satisfying a task requirement by extracting an answer, comparing values, modifying page state.}
    \label{fig:navigational-actions}
\end{figure}

\begin{figure}[t]
    \centering
    \includegraphics[width=0.8\columnwidth]{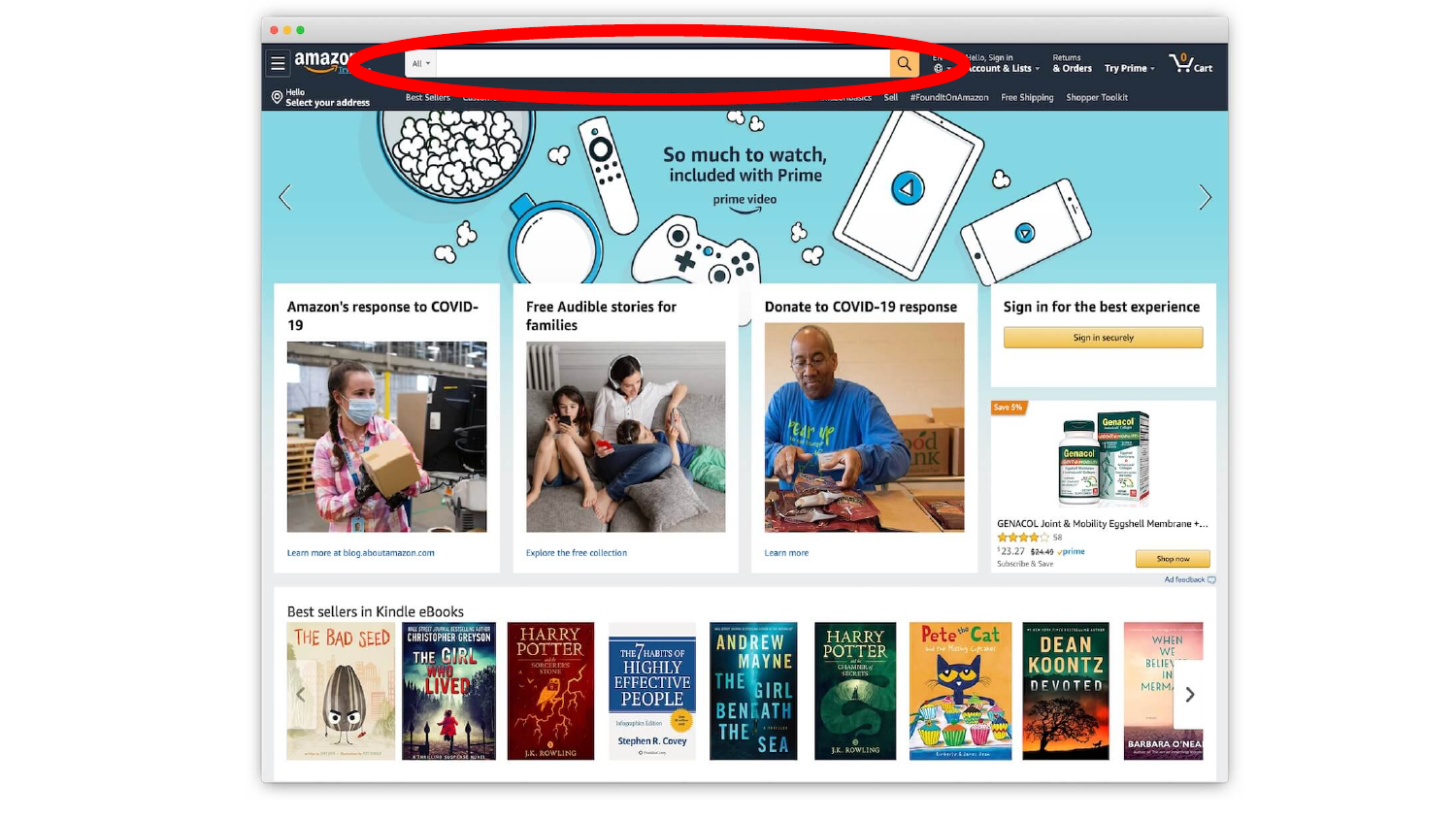}
    \tightcaption{Web state input at one step of a task. Red circle marks the search bar, the only
element load-bearing for the next action (typing the query). The
rest of the input (product recommendations,
images, sidebar links, etc.) is not needed.}
    \label{fig:excessive-data}
\end{figure}

\para{Most steps are navigational.}
Most web agent tasks involve read-only information retrieval (a product price, a paper's citation count, a course's prerequisites), with the answer typically concentrated on one or a few destination pages. Reaching those pages involves searching, filtering, paginating, clicking through results, and so on. However, these steps are primarily navigational -- steps used to move between pages rather than to directly satisfy a task requirement -- rather than load-bearing reasoning. Figure~\ref{fig:navigational-actions} illustrates this, showing that 66.7\% of steps are purely navigational for the median task in our benchmarks. Further, on a given site, this navigation often follows query-invariant patterns: every Amazon shopping task involves search-then-product-page navigation, every arXiv lookup resolves an identifier to an abstract page, every GitHub task locates a repository by owner and name. An agent that knows the pattern can thus reach the destination in a single direct fetch rather than rediscovering it via a multi-step ReAct loop, with only the query-specific identifiers substituted in.

\begin{figure}[t]
    \centering
    \begin{minipage}{0.48\columnwidth}
        \centering
        \includegraphics[width=\linewidth]{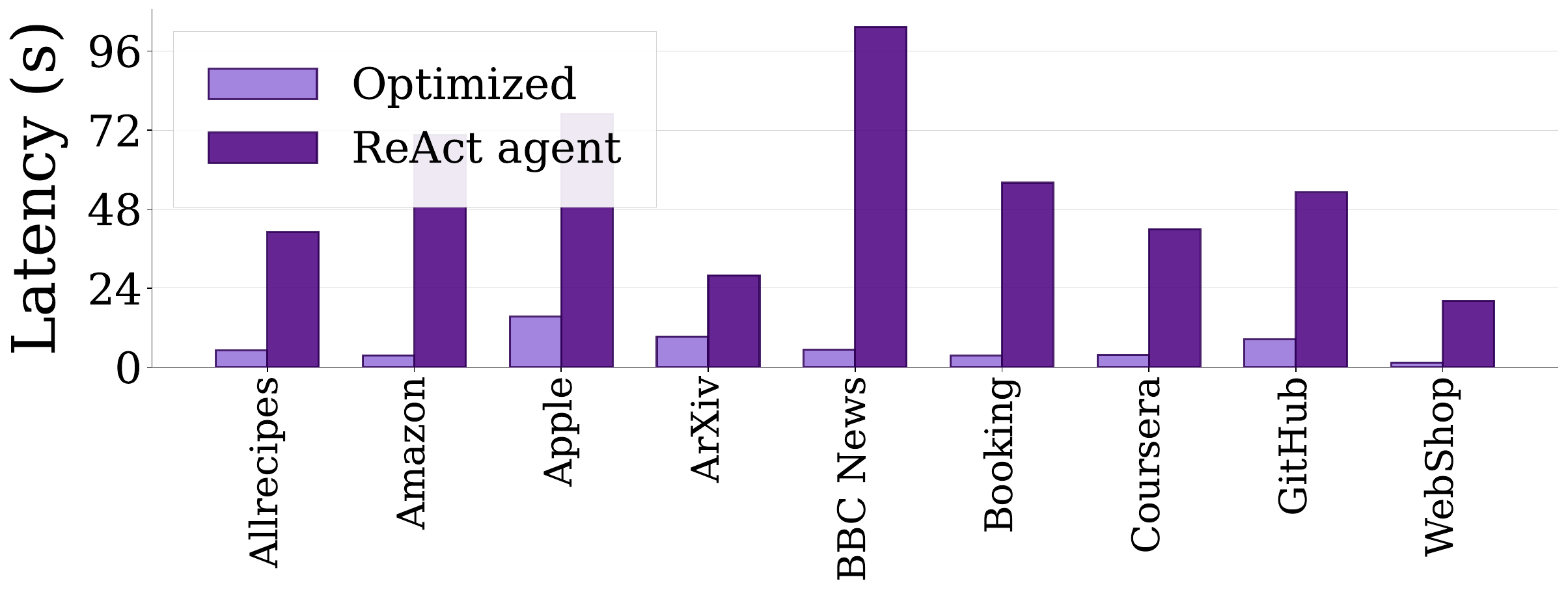}
        \tightcaption{Latencies of handcrafted optimized programs.}
        \label{fig:hand-optimized-latency}
    \end{minipage}
    \hfill
    \begin{minipage}{0.48\columnwidth}
        \centering
        \includegraphics[width=\linewidth]{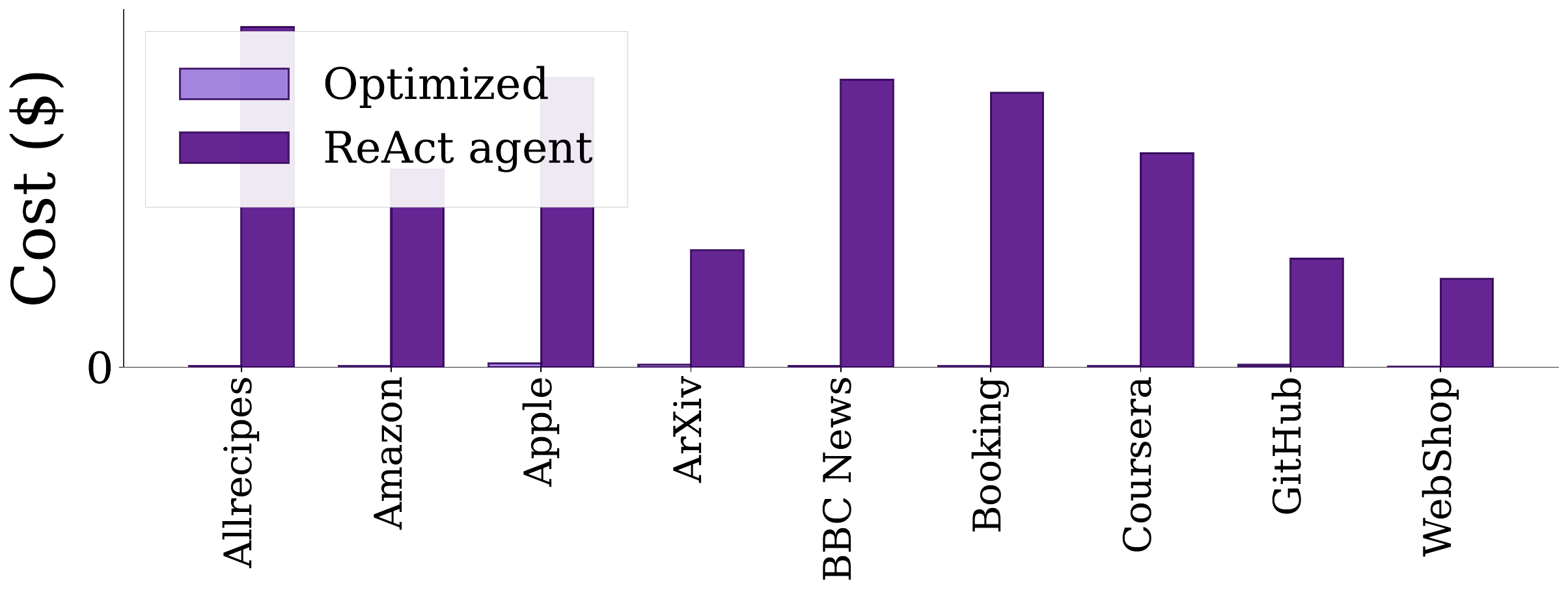}
        \tightcaption{Costs of handcrafted optimized programs.}
        \label{fig:hand-optimized-cost}
    \end{minipage}
\end{figure}

\begin{figure}[t]
    \centering
    \includegraphics[width=0.8\columnwidth]{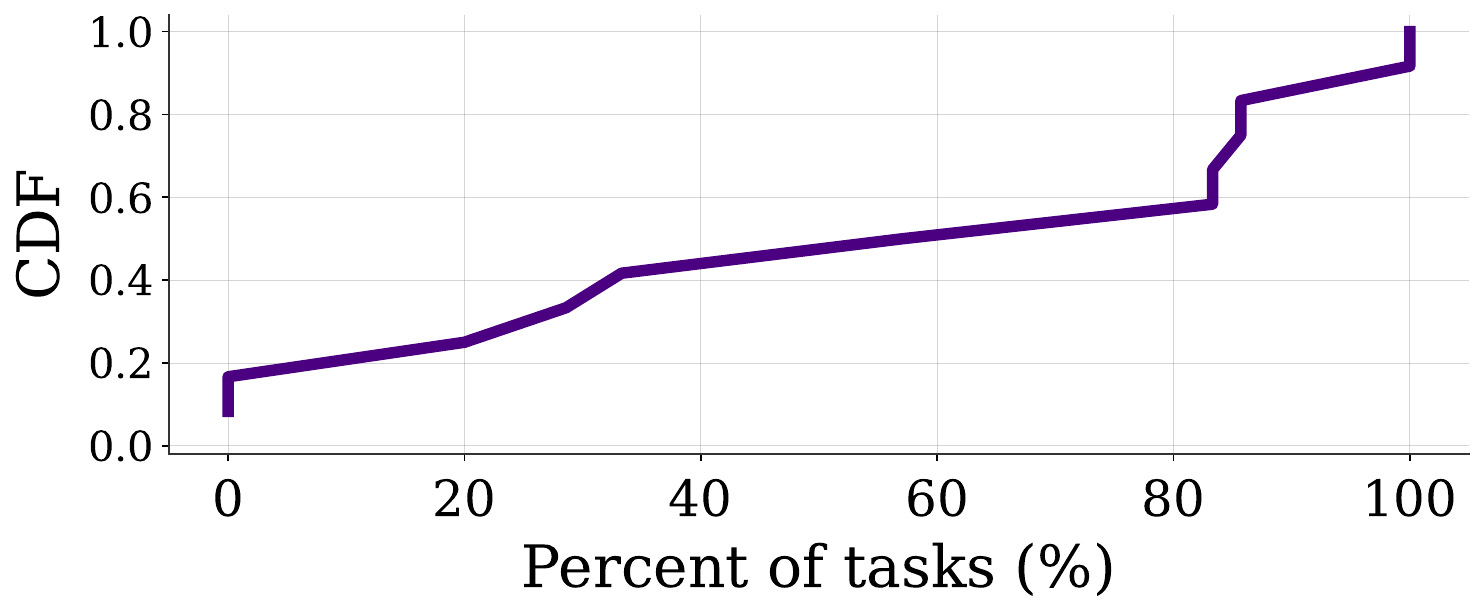}
    \tightcaption{Fraction of tasks per site solvable via HTTP-only execution. Many sites support a large percentage of tasks through direct retrieval without browser interaction.}
    \label{fig:http-coverage}
\end{figure}

\para{Many pages do not need a browser.}
For many sites, the content embedding a task's answer is rendered on the server and arrives complete in the HTTP response body (e.g., Amazon product pages, arXiv abstracts, GitHub repository views). In these cases, a plain HTTP fetch returns what the browser would have displayed, without the high overheads of scripting and rendering. Figure~\ref{fig:http-coverage} confirms that this is common: for 55.8\% of tasks in our benchmark, the content necessary to resolve the query is fully accessible via HTTP fetch alone, without invoking the browser. Browser execution remains necessary only when client-side JavaScript produces load-bearing content (e.g., interactive maps, dynamically-revealed product variants), and even then, these overheads are warranted only for the specific steps that depend on the dynamically-generated content. Skipping even a subset of browser executions can eliminate substantial overhead.

\begin{figure}[t]
    \centering
    \includegraphics[width=0.8\columnwidth]{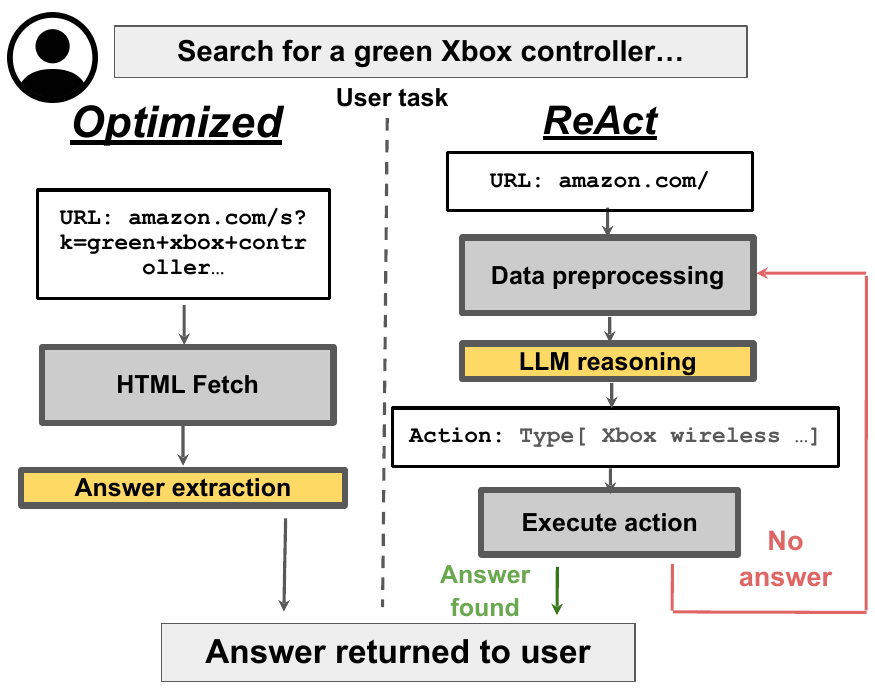}
    \tightcaption{Comparison of a hand-optimized program (left) and the
corresponding off-the-shelf ReAct trajectory (right) for an
Amazon retrieval task. }
\vspace{4pt}
    \label{fig:optimized-web-program}
\end{figure}

 \begin{table}[t]
    \centering
    \small
    \begin{tabular}{lcc}
        \toprule
        \textbf{Model} & \textbf{Denoised environment} & \textbf{Full environment} \\
        \midrule
        Qwen2.5-14B & 42.7\%  & 26.1\%  \\
        GPT-4o &  45.7\% & 45.0\% \\
        \bottomrule
    \end{tabular}
    \vspace{8pt}
    \tightcaption{Effect of page noise on extraction accuracy. Qwen2.5-14B matches GPT-4o on small, denoised page content embedding only the necessary information, but degrades sharply on full pages while GPT-4o remains comparatively robust.}
        \label{t:accuracy-vs-noise}
\end{table}

\begin{figure}[t]
    \centering
    \includegraphics[width=0.8\columnwidth]{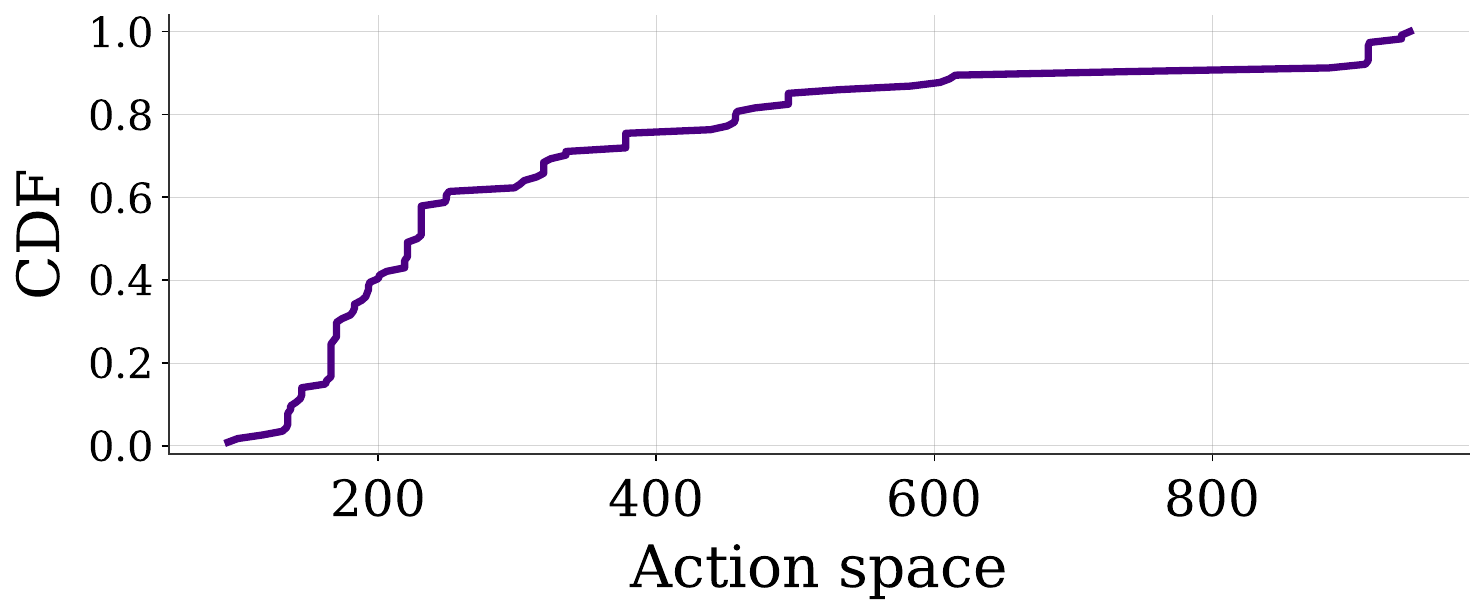}
    \tightcaption{CDF of navigational elements (links, buttons, form
inputs) per page across ReAct trajectories. The median page
presents 212+ navigational elements, each leading to a different
page. }
    \label{fig:nav-elements-challenge1}
\end{figure}

\para{Most steps do not need a frontier model.}
A typical web page presents hundreds of potential actions in the form of interactive elements alongside ads, recommendation widgets, navigation chrome, and related-content sidebars (Figure~\ref{fig:nav-elements-challenge1}). However, the content directly relevant to a given step (e.g., a price, an abstract, a commit count) often occupies only a small fraction of the page's DOM content; Figure~\ref{fig:excessive-data} shows an illustrative example. Powerful frontier models are necessary to reason over this combination of useful content and noise, but the underlying extraction -- i.e., locating and returning the page content relevant to the step -- is not inherently hard. The relevant region is not only small but predictably located: a product's price routinely appears in the same place across product pages, an ArXiv abstract appears in the same place across paper pages, etc. Once that region is isolated, extraction reduces to a structured task within a smaller model's competence. 
Table~\ref{t:accuracy-vs-noise} confirms this empirically: when presented only with the region of a page embedding the answer (i.e., `denoised'), non-frontier models nearly match frontier-model accuracy; on full pages, however, non-frontier accuracy plummets while frontier models remain capable.

\para{Opportunity.}
To quantify the savings these three properties enable, we hand-engineer specialized programs for nine representative tasks in the WebVoyager benchmark. Each program implements the same recipe illustrated in Figure~\ref{fig:optimized-web-program}: a per-site URL template constructs the target page URL, an HTTP request fetches the page, the HTML is filtered to the answer-bearing region encoded by the site's expected answer structure, and a non-frontier model (i.e., Qwen2.5-14B-Instruct rather than GPT-4o/GPT-5) extracts the answer. As Figures~\ref{fig:hand-optimized-latency} and \ref{fig:hand-optimized-cost} shows, these fast-path programs run 66.7-94.9\% faster 17.7-100.7$\times$ lower costs) than the off-the-shelf ReAct agent, while preserving accuracy.

\begin{figure}[t]
    \centering
    \includegraphics[width=0.8\columnwidth]{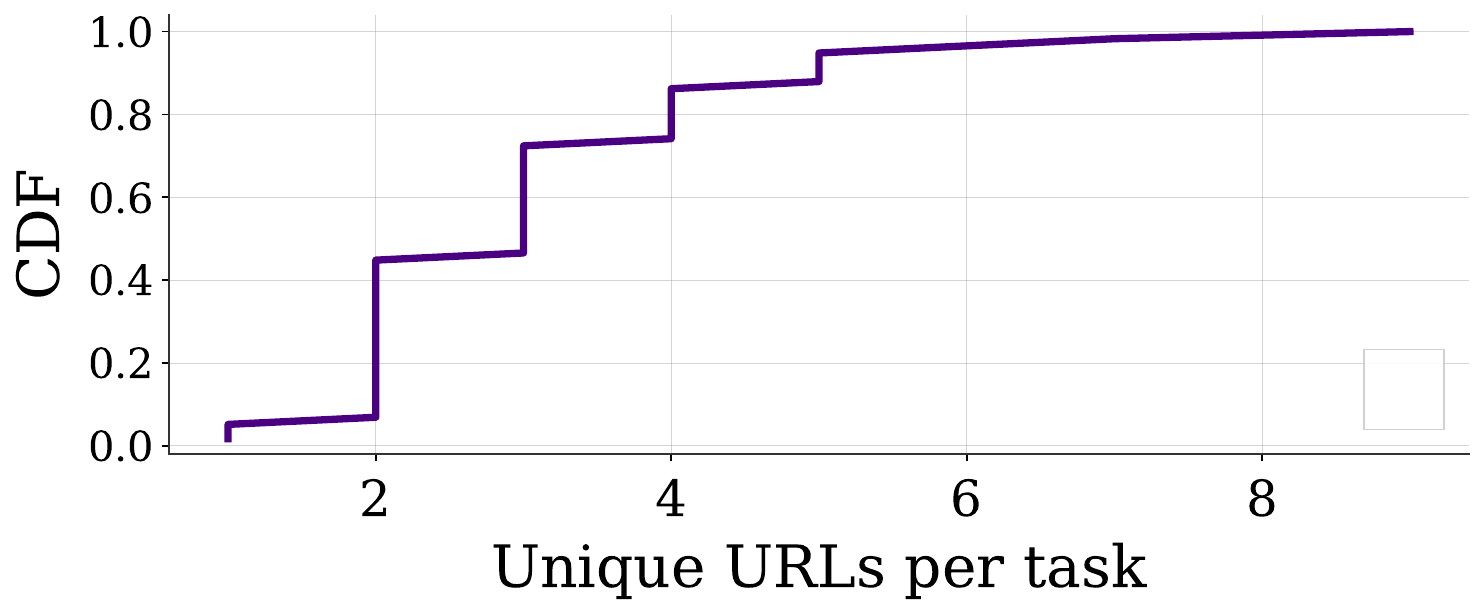}
    \tightcaption{Number of unique URLs visited by each task. Tasks tend to take very different trajectories. Note that multiple steps often occur within each URL (e.g., scrolling) }
    \label{fig:unique-per-task-url-cdf-challenge1}
\end{figure}

\begin{figure}[t]
    \centering
    \includegraphics[width=0.8\columnwidth]{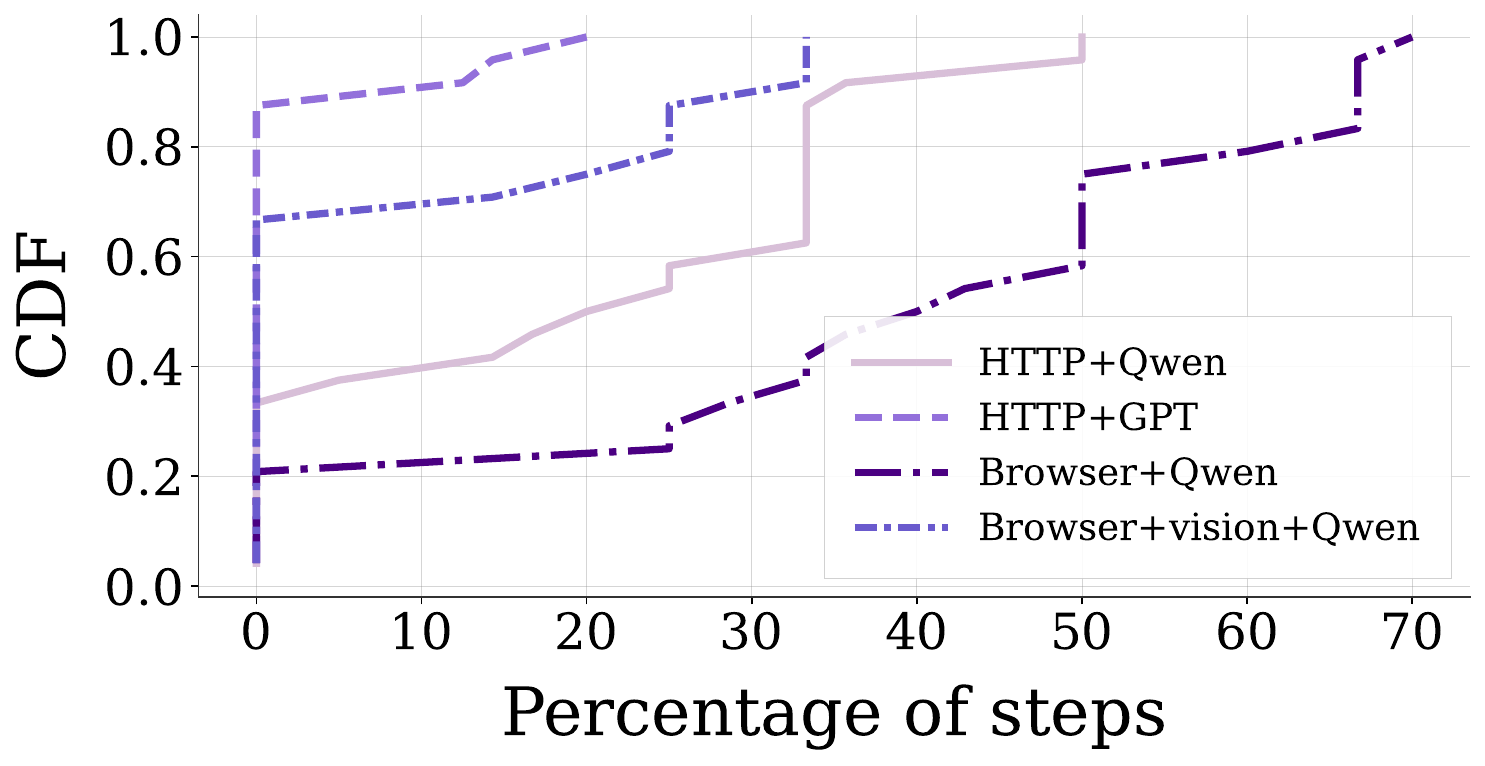}
    \tightcaption{CDF over tasks of the percentage of steps requiring each compute tier as the weakest tier sufficient to match the full ReAct agent. Tiers combine page-fetch (HTTP, browser, or browser w/ screenshot) with a model (Qwen2.5-14B or GPT-4o).}
    \label{fig:minimum-tier-required-challenge2}
\end{figure}

\subsection{Challenges}
\label{sec:chal}

Automatically realizing the promise of specialization above requires overcoming two key challenges, which we describe in turn. The cost of getting either wrong is severe: naively substituting cheap components (a non-frontier model and HTTP fetch) into the off-the-shelf ReAct loop, without the URL templates and answer schemas that hand-engineered programs encode, drops average task success by 60\% since errors compound across the trajectory and a single wrong step can invalidate everything downstream.


\para{C1: Determining \emph{what} can be specialized.}
Before execution begins, the agent must determine which parts of an incoming task support specialization, i.e., which trajectories can be replaced by direct URL fetches, which pages can be served without a browser, and which extraction steps can use a non-frontier model. This determination is hard for two reinforcing reasons. First, the signal available at query time is thin. Task text describes user intent (``find the cheapest blue headphones'') rather than the trajectory pattern that resolves it, and even on a single site, different tasks visit different pages through different navigation sequences. Second, recognition itself has a cost ceiling: if identifying reuse opportunities requires running a frontier model over the full site or comparing against a large trajectory database, the cost of recognition negates the savings that the specialization would yield. The challenge is thus how to encode reusable site structure ahead of time and match incoming tasks against it efficiently at runtime.


\para{C2: Determining \emph{how} to specialize across steps.}
Even when specialization applies, the minimum-cost \emph{tier} -- a combination of retrieval, rendering, and extraction model -- sufficient to match the default agent varies per step within a single trajectory. Figure~\ref{fig:minimum-tier-required-challenge2} highlights this by showing the distribution of minimum-tier support needed across task steps. As shown, some steps are resolved by the cheapest tier (HTTP fetch with a non-frontier model), others require browser rendering to access client-side content, others require a frontier model for extraction the small model cannot perform, and some require both. Importantly, the variation is \emph{per step}, not just per task; even tasks amenable to specialization mix step types within their trajectories. Choosing the wrong tier at any step drops accuracy substantially. For instance, a small model handed a raw DOM with a complex extraction task might return the wrong field, while an HTTP fetch of a JavaScript-dependent page will return a content stub without the necessary information. The challenge is to determine the minimum sufficient tier per step without paying the cost of running the default agent to find out.

\begin{figure*}[t]
    \centering
    \includegraphics[width=0.8\textwidth]{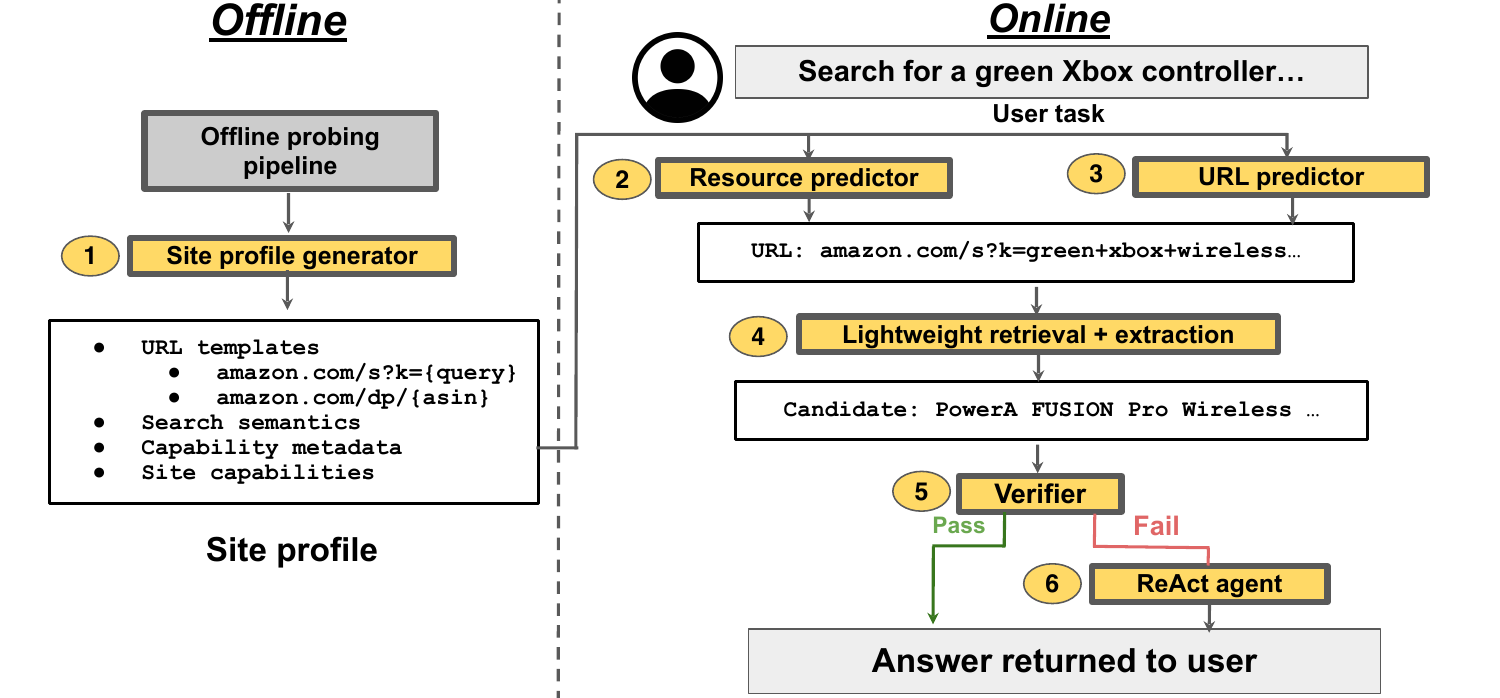}
    \tightcaption{\sys end-to-end. Offline (left): a profiling pipeline
    encodes per-site URL templates, search behavior, and extraction
    schemas into a reusable site profile. Online (right): tasks attempt
    direct addressing through template-driven URL synthesis, plain HTTP
    fetch, and lightweight extraction. Output is verified before
    commitment; on rejection, the system cascades to a heavier
    execution tier resuming at the URL the fast path reached.}
    \label{fig:design}
\end{figure*}

\section{Design of \sys{}}

\subsection{Overview}
To address these challenges, we introduce \sys{}, a framework that accelerates web agents through speculative lightweight execution while preserving robustness through verification and selective escalation only when needed. Figure~\ref{fig:design} illustrates \sys{}'s end-to-end operation. The system rests on two complementary observations about purpose-built websites. First, website structure is sufficiently stable that reusable navigation patterns, search semantics, and capability characteristics can be extracted offline and amortized across future queries. Second, most read-only web tasks do not require full iterative browser interaction — they can often be resolved through direct retrieval and lightweight extraction once the correct URL and execution environment are known. \sys{} therefore combines offline structural profiling with an online speculative cascade: offline profiling provides reusable trajectory priors, while runtime verification and fallback preserve correctness when speculation fails. These two mechanisms jointly address C1 and C2; offline patterns alone are insufficient because websites contain unpredictable edge cases, while runtime adaptation alone lacks the structural priors needed to make speculative execution efficient.

\sys{} interposes between the user and an underlying ReAct agent. Before runtime, an offline profiling stage \textcircled{1} constructs per-site profiles encoding URL templates, search semantics, pagination behavior, query-construction guidance, and execution-relevant capability metadata. For each incoming task, a capability and resource predictor \textcircled{2} determines the minimum execution environment and reasoning resources likely required, while a structure predictor \textcircled{3} attempts to synthesize a high-probability retrieval URL using the site profile. The system then performs lightweight retrieval and extraction \textcircled{4}, fetching the page through the cheapest supported mechanism and running schema-guided extraction over cleaned HTML. A verifier \textcircled{5} checks whether the speculative result is consistent with the task: if verification succeeds, the answer is returned immediately; otherwise, execution escalates to the full ReAct agent \textcircled{6}, seeded with the speculative URL as a warm start to preserve navigational progress. Tasks that match no reliable template, as well as state-mutating tasks outside the read-only scope (\S\ref{sec:background}), bypass speculative execution and route directly to the default ReAct agent. We describe offline profiling pipeline in \S\ref{sec:design:offline}, and runtime cascade in \S\ref{sec:design:runtime}.

\sys's speculation primitive admits two deployment policies. In accelerate mode, a verified fast path commits immediately and the saved compute is returned to the user as reduced latency and cost. In aggregate mode, the saved budget is reinvested into additional speculative trials within the baseline's single-execution cost envelope; the verifier ranks the resulting candidates and selects the best. Both modes share the same offline profiles, runtime URL synthesis, and verifier; they differ only in what is done with the verifier's output. Accelerate mode treats verification as a commit gate; aggregate mode treats it as a ranking function. 

\subsection{Offline Profiling Pipeline}
\label{sec:design:offline}

The offline pipeline produces a per-site profile that captures
reusable trajectory structure. Site URL patterns and structural
conventions are stable across queries, e.g., Amazon products are
addressable by a stable product identifier (Amazon Standard
ID), arXiv papers by identifier,
and GitHub repositories by \texttt{owner/repo}. The profile can
thus be built once per site and reused across subsequent
tasks. This stability is what makes C1's prediction problem
tractable despite thin task-time signal: rather than inferring
reusable patterns at runtime from task text and a brief view of
the site, the system matches each task against an already-built
site profile.

\para{What a profile contains.}
A site profile encodes the reusable structural information \sys{} needs to specialize execution. 
The profile contains rich search semantics, pagination behavior, capability measurements, extraction schemas, verification metadata, and query-construction guidance. Each profile has four main components. First, typed URL templates capture reusable navigation patterns (direct identifier lookup, filtered search, sorted retrieval, paginated traversal); each template contains parameterized URL patterns and regex constraints that map natural-language task descriptions into site-valid parameter values. Second, search semantics encode how a site constructs queries, exposes filters, handles sorting and pagination, signals empty results, and maps structured attributes (e.g., price ranges, categories, authors, difficulty levels) into URL parameters. Third, answer schemas specify the expected structure and types of valid responses, enabling both focused extraction and lightweight runtime verification. Fourth, capability metadata records execution-relevant properties measured during probing: JavaScript dependence, HTTP accessibility, bot-detection behavior, rendering constraints, response latency, and content-density statistics.

Profiles also store compact query-construction notes and representative search queries generated offline, guiding runtime URL synthesis toward high-coverage search paths without overwhelming downstream models with large template spaces. The goal is not to memorize every possible trajectory, but to capture the small set of structurally common navigation patterns that cover most realistic tasks.

\para{How profiles are constructed.}
Profiles are generated by an automated offline probing pipeline that runs once per site. The pipeline characterizes the site's execution environment through lightweight HTTP probing, measuring response latency, visible-content density, HTTP accessibility, and bot-detection behavior; when HTTP retrieval returns sparse or incomplete content, it escalates to a headless browser to recover the fully rendered page. Using the best available HTML, it then discovers search endpoints by analyzing form structure and common search-URL conventions, identifies pagination behavior by traversing result pages, and validates candidate URL templates against real site behavior. Rather than enumerating all possible trajectories, the prober strategically searches for reusable structural patterns that maximize coverage across common task classes while keeping the template space compact for efficient runtime matching. This reflects an empirical property of purpose-built sites: although URLs and content vary enormously across queries, the underlying navigation mechanisms are highly similar — product sites repeatedly expose search, filter, sort, and pagination flows; literature sites repeatedly expose identifier lookup and query retrieval flows; repository sites repeatedly expose repository, issue, and commit navigation flows.

The pipeline further enriches each profile using a local LLM that analyzes discovered search endpoints, validated URL patterns, and representative task descriptions to synthesize query-construction guidance and representative search queries. These enrichment stages help \sys{} learn which task attributes are structurally useful for retrieval on a given site and which are likely to hurt coverage or overconstrain search. For example, shopping sites benefit from including product attributes (color, material, device type), while review-count thresholds or highly specific descriptive modifiers often reduce retrieval quality. The resulting profiles therefore encode not only URL structure but also higher-level retrieval strategies specialized to each site's intended purpose.

\para{Handling site changes.} Site profiles persist for long periods without growing stale. While dynamic content changes continuously (Amazon rotates products hourly, arXiv publishes new papers daily), structural scaffolding -- search endpoints, URL parameter conventions, pagination behavior, navigation flows, and answer-bearing page layouts -- evolves much more slowly because it is tied to the site's core purpose, and the workflows that purpose supports remain stable over time. \sys exploits this asymmetry: profiles encode the slow-changing structural layer while leaving dynamic content to be resolved online at runtime.

Structural drift is therefore uncommon but not ignored.
\sys{} continuously validates profile correctness during runtime
execution through schema verification, HTTP status monitoring,
and content-consistency checks. Missing expected fields,
repeated verification failures, unexpected page structures, or
persistent HTTP errors (e.g., 404 responses on previously valid
templates) are treated as indicators that a profile may have
become stale. When such signals exceed a threshold, \sys{}
automatically re-invokes the offline probing pipeline to
regenerate the affected portions of the profile. Because
regeneration is amortized across all future tasks on the site,
occasional reprofiling incurs negligible overhead relative to
sustained query volume.

\para{URL synthesis.}
URL synthesis is the runtime mechanism that maps a task description to a URL using the profile's templates. It has two steps: selecting the right template (search vs direct lookup, with which filters and sort orders applied) and extracting parameter values to slot in (query keywords, identifiers, filter values). On Amazon, ``find the cheapest blue headphones under \$100'' selects the price-range search template with price-ascending sort, extracts ``blue headphones'' as the query and ``\$100'' as the maximum price (converted to cents per the template's parameter description), and produces ``{\nolinkurl{amazon.com/s?k=blue+headphones\&rh=p\_36\%3A0-10000\&\\s=price-asc-rank}}''. The synthesized URL collapses several steps the default agent would otherwise execute through multi-step navigation (search, filter, and sort).

URL parameters codify much of a site's interactive surface (search keywords, filter values, sort orders, navigational pages), so a synthesizer that handles parameters correctly can reach most of the destinations the default agent would. \sys{} performs template selection and parameter extraction with a non-frontier model, supported by per-template regex patterns in the profile: the model proposes candidate values from the task text, and the regexes sanity-check them against the site's expected vocabulary before URL construction. The regexes are the bridge between natural-language tasks and structured URL parameters, reducing the model's job to pick the right template and surfacing the right value spans.

A non-frontier model suffices due to the structured nature of purpose-built sites. Although natural-language queries are arbitrary in principle, the \emph{semantically reasonable} queries for a given site are much narrower than the unrestricted space — nobody asks to read a research paper on Amazon or compare iPhone prices on arXiv. The site's purpose constrains the queries the synthesizer must handle to a set whose parameter types are encoded in the profile, so the model never needs to handle truly arbitrary intent. A general-purpose URL constructor that handles arbitrary tasks across arbitrary sites would benefit from a frontier model; a per-site synthesizer operating within its site's intended vocabulary does not.

\begin{figure}[t]
    \centering
    \includegraphics[width=0.99\columnwidth]{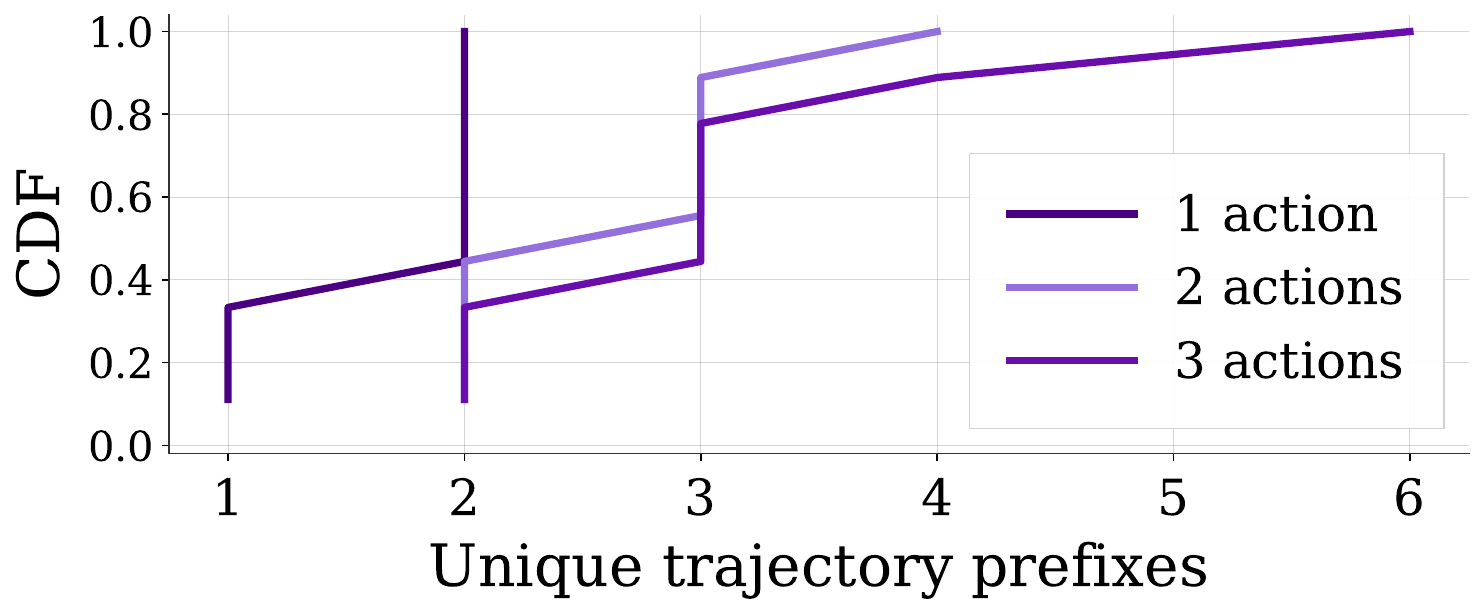}
    \tightcaption{Distribution of the number of unique action-type prefixes across tasks on a site, for the first 1, 2, and 3 actions of each task. Prefixes consider action types (e.g., click, type), ignoring parameters like element identifiers or extracted content.}
    \label{fig:trajectory-prefixes}
\end{figure}

\subsection{Runtime Speculation}
\label{sec:design:runtime}

When a task arrives, \sys matches the task against the profile's typed templates through a hybrid process: a lightweight LLM performs intent classification to select a candidate template class (e.g., direct identifier lookup, filtered search, paginated retrieval), while profile-defined regex constraints and parameter validators ground extracted slot values into site-valid parameters. The LLM thus handles semantic interpretation while typed parameter constraints enforce structural validity and prevent free-form generation. For example, ``find the abstract of arXiv paper 2401.12345'' matches the direct-identifier template with slot \texttt{2401.12345}, while ``find the cheapest blue headphones on Amazon'' matches the filtered search template with query, sort-order, and optional price slots. This combination is more robust than strict string or regex-only matching, and more reliable and cheaper than unconstrained frontier-model trajectory planning.
If no template matches with sufficient confidence, the task routes directly to the default ReAct agent. Otherwise, \sys determines the minimum resources the task requires from the profile, executes the speculative fast path, verifies the output, and escalates resources on rejection.



Resource determination operates along three axes that govern how
a task is executed:
\begin{sloppypar}
\begin{itemize}
    \item \textbf{Page acquisition.} Either direct URL fetch
    (the page is reachable by URL synthesis from the profile
    template) or multi-step navigation (the ReAct loop). The
    profile dictates which is applicable for the task class.
    \item \textbf{Page rendering.} Either HTTP-only fetch or
    browser-driven rendering. The profile encodes which pages
    on a site require browser rendering (client-side JavaScript,
    authenticated state); pages flagged as HTTP-reachable use
    HTTP, the rest use the browser.
    \item \textbf{Reasoning model.} Either a non-frontier
    extraction model or a frontier model. The profile's answer
    schema and the cleaned page content together determine
    whether the non-frontier model can extract reliably; otherwise a frontier model is used.
\end{itemize}
\end{sloppypar}

The cheapest combination the profile justifies is tried first.
On verification failure, \sys{} escalates the axis the verifier
signals is at fault: a rejection on a missing-but-expected field
escalates the reasoning model, a rejection on an empty or blocked
page escalates rendering to the browser, and persistent failure
escalates page acquisition to full ReAct. Escalation is not a
strict hierarchy; different tasks exercise different axes. Some
need browser rendering with a non-frontier model, others need
HTTP with a frontier model, and many need only the cheapest
combination on all three. The verifier is the feedback signal
that drives axis-specific escalation rather than blind progression
through fixed tiers.


This feedback-driven escalation rests on two empirical properties of purpose-built websites: trajectory prefixes are highly predictable, making failed speculation useful, and verification is highly constrained, making it cheap to evaluate.


\para{Trajectories share structural prefixes.} On purpose-built sites, queries with different end goals typically begin with identical navigational steps (e.g., search, filter, sort) (Figure \ref{fig:trajectory-prefixes}). This split aligns with \sys{}'s offline/online design: the offline profile captures the shared prefix, while runtime synthesis executes the task-specific tail. Because the speculative fast path excels at navigating these common prefixes, even a trajectory that ultimately fails usually reaches a useful intermediate URL. A more expensive recovery policy can then inherit this progress rather than starting from scratch.

\para{Verification is structurally cheap.}
Checking whether a read-only trajectory succeeded is inexpensive for two reasons. First, the site's profile provides a strong structural prior: it defines the valid navigation templates, reducing the verification problem to checking alignment between the expected task structure and the current state. Second, web state compresses well: a short textual summary of the retrieved content, combined with the URL, is usually sufficient to confirm or reject the outcome without relying on expensive full-page DOM processing. We empirically validate this efficiency against a frontier-model verifier in \S\ref{sev:eval}.


\para{HTML cleaning is a prerequisite.}
As shown in \S\ref{sec:background}, non-frontier models perform substantially better when input is condensed to task-relevant regions. \sys therefore filters each fetched page against the profile's answer schema before extraction or verification, retaining content matching expected response structure (titles, prices, abstracts) and discarding the rest. This serves both stages: the extraction model operates on focused evidence, and the verifier receives a smaller input that keeps its cost cheap.


\para{Two-stage verification.}
The verifier spans the failure-mode spectrum identified in C2 at
low marginal cost. The first stage is a near-zero-cost schema
check: the candidate answer must be non-empty, type-compatible
with the task, and within the expected value range encoded in
the profile. This stage catches obvious failures (bot-detection
pages, empty results, type-mismatched answers) without
an LLM. The second stage runs only on answers that pass the
schema check: a non-frontier semantic judge presented with a
compressed state summary (task description, current URL, cleaned
page summary, candidate answer) decides whether the trajectory
is consistent with the task. Both stages reuse machinery already
provisioned for the fast path. Crucially, the verifier must
be cheap: if verifying the fast path required a frontier
model over the full DOM, the cost of verification would negate
the savings of the fast path. The two-stage design keeps
verification within the per-task cost envelope the fast path
establishes and false positives waste only a retry.

\para{Warm-start cascade.}
On verification rejection, \sys{} retries the task with
escalated resources, inheriting the URL the prior attempt
reached. The cascade preserves the navigational progress the
cheaper attempt made correctly (per the prefix-sharing
observation above), discarding only what the verifier rejected.
Only the URL is propagated: the rejected attempt's extracted
answer and reasoning trace are not, and the escalated attempt
performs its own independent reasoning from the warm-start page.
The URL itself is monotone progress in the sense that escalation
never starts from a position worse than where the cheaper attempt
reached. ReAct's per-step adaptation is designed to handle
exactly this kind of recovery, so a default-agent
cascade resuming at a warm-start URL behaves as a normal ReAct
trajectory starting from that URL rather than from the task's
initial page.

Incorrect warm starts, although infrequent, are also relatively inexpensive to recover from. Because intermediate navigation on read-only tasks is scaffolding rather than load-bearing reasoning (\S\ref{sec:background}), landing on the wrong URL does not require undoing irreversible state mutations or reasoning dependencies -- the agent only needs to redirect navigation toward a different destination page. ReAct's iterative-recovery design handles this naturally, so recovering from an incorrect warm-start URL behaves like any other mid-trajectory navigation adjustment. 

\subsection{Query Support and Deployment Discussion}

\begin{sloppypar}
\sys targets read-dominant web workloads, including search, comparison, structured extraction, and exploratory navigation. These workloads constitute a large fraction of modern  deployments, including retrieval-augmented assistants, deep-research systems, shopping assistants, and enterprise information-gathering pipelines. In these settings, execution cost is dominated not by the final interaction itself, but by exploratory navigation and repeated frontier-model inference over intermediate pages.
\end{sloppypar}

Stateful actions such as purchases, form submissions, or account modifications impose substantially stronger correctness and safety requirements because incorrect execution may produce irreversible side effects. Supporting such workflows robustly requires infrastructure for authentication management, transactional safety, and side-effect validation, which we leave to future work. Nonetheless, workflows that ultimately perform state mutations are typically dominated by long read-only prefixes involving search, filtering, comparison, and navigation. \sys accelerates this expensive exploratory phase while delegating irreversible transactional actions to conventional browser-agent execution.

A second design goal is broad applicability across arbitrarily-worded tasks and heterogeneous websites. Rather than hardcoding a fixed collection of domains or task schemas, lightweight language models map free-form task descriptions onto profile-encoded patterns through typed template matching and parameter extraction. As more sites are profiled, the system naturally expands coverage without requiring changes to the core runtime architecture.

\section{Implementation}
\sys{} is implemented in Python as a multi-tier speculative
execution layer atop existing ReAct-based web agents. The
offline pipeline performs one-time per-site analysis through
HTTP probing with headless-browser escalation, discovering
search endpoints, URL templates, pagination behavior, rendering
requirements, and bot-detection characteristics; a local
Qwen2.5-14B-Instruct model served via vLLM~\cite{vllm} enriches each profile
with query-construction guidance and representative search
templates. At runtime, the URL predictor synthesizes destination
URLs through regex extraction, typed URL templates, and
Qwen-based slot filling, while the resource predictor selects
the minimum execution environment and model from the task,
retrieved content, and offline capability metadata. Tasks are
then mapped to one of four execution tiers ranging from direct
HTTP+Qwen retrieval to full ReAct; lightweight tiers perform
schema-guided HTML cleaning and extraction with Qwen2.5-14B, and
higher tiers invoke browser navigation or off-the-shelf ReAct
agents (WebVoyager, BrowserUse, AgentOccam) configured with
GPT-4o, consistent with prior web-agent evaluations and benchmark configurations. A Qwen-based verifier guards speculative outputs and
cascades to heavier tiers on rejection, preserving navigational
progress through warm-start URLs produced by earlier stages.

\section{Evaluation}
\label{sev:eval}

\begin{figure*}[t]
    \centering

    \begin{minipage}{0.32\textwidth}
        \centering
        \includegraphics[width=\linewidth]{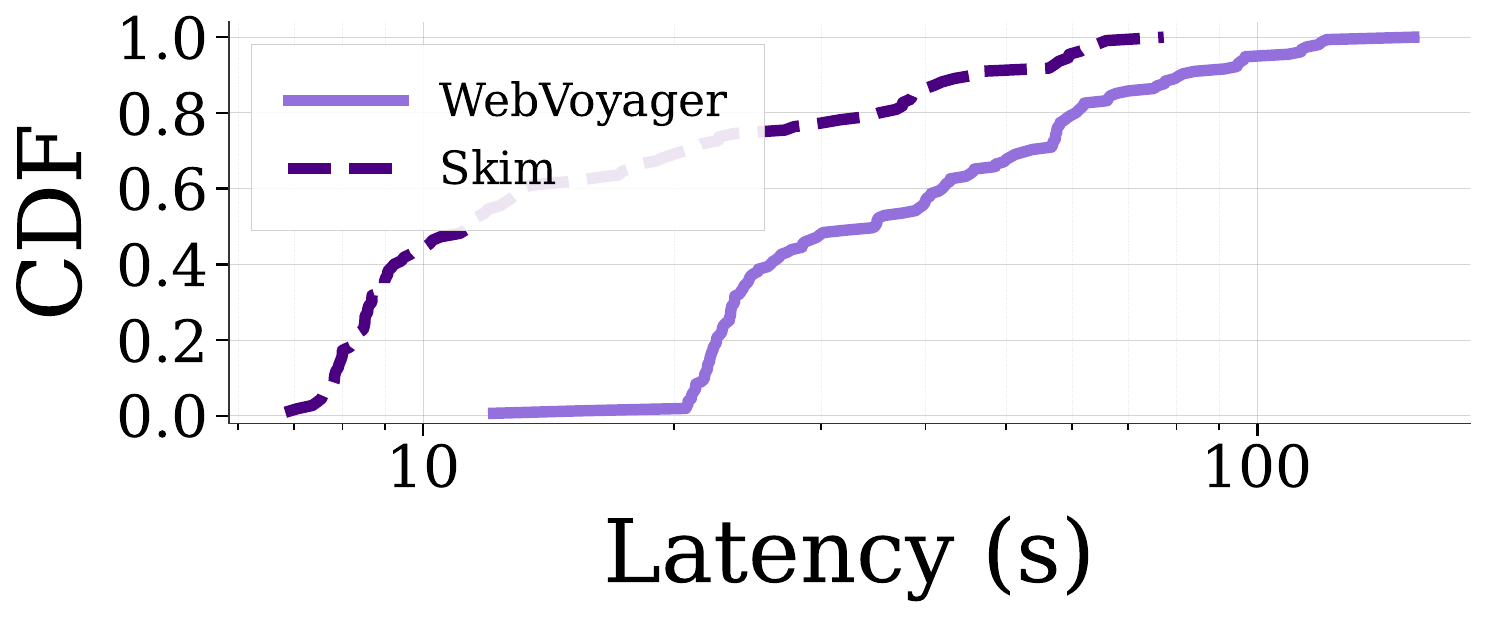}
    \end{minipage}
    \hfill
    \begin{minipage}{0.32\textwidth}
        \centering
        \includegraphics[width=\linewidth]{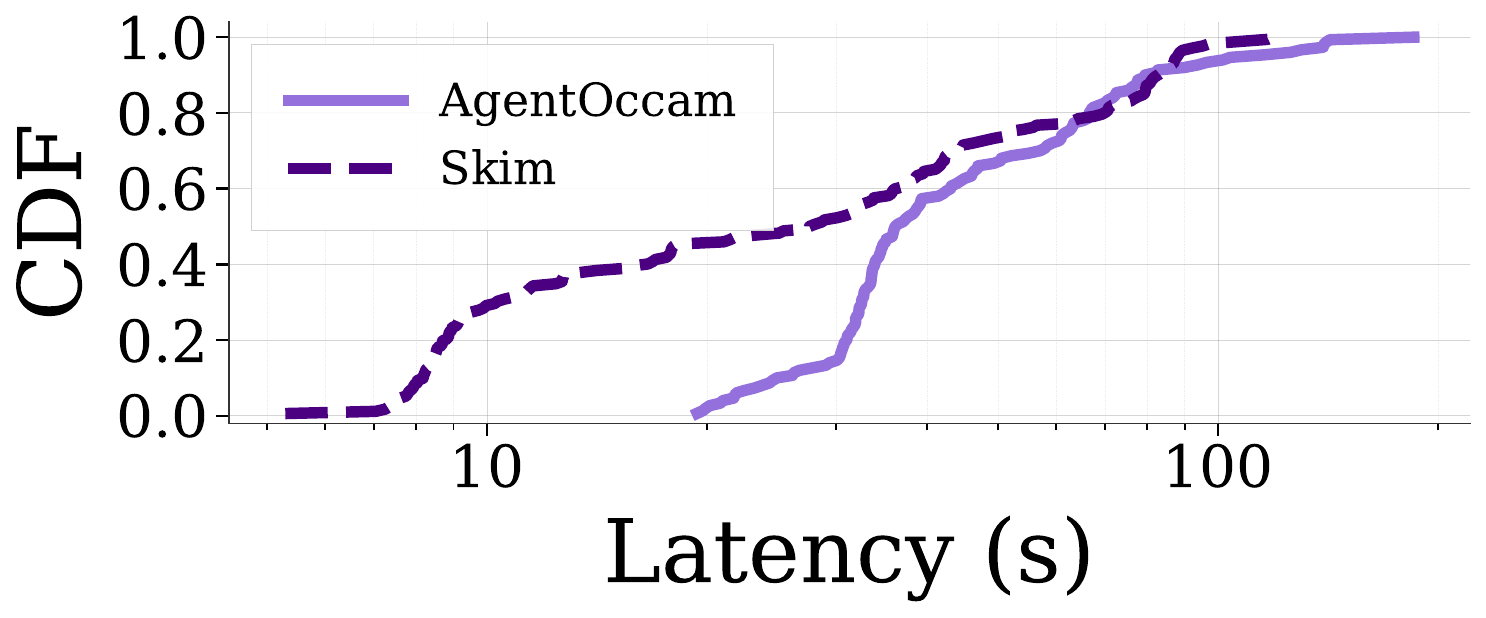}
    \end{minipage}
    \hfill
    \begin{minipage}{0.32\textwidth}
        \centering
        \includegraphics[width=\linewidth]{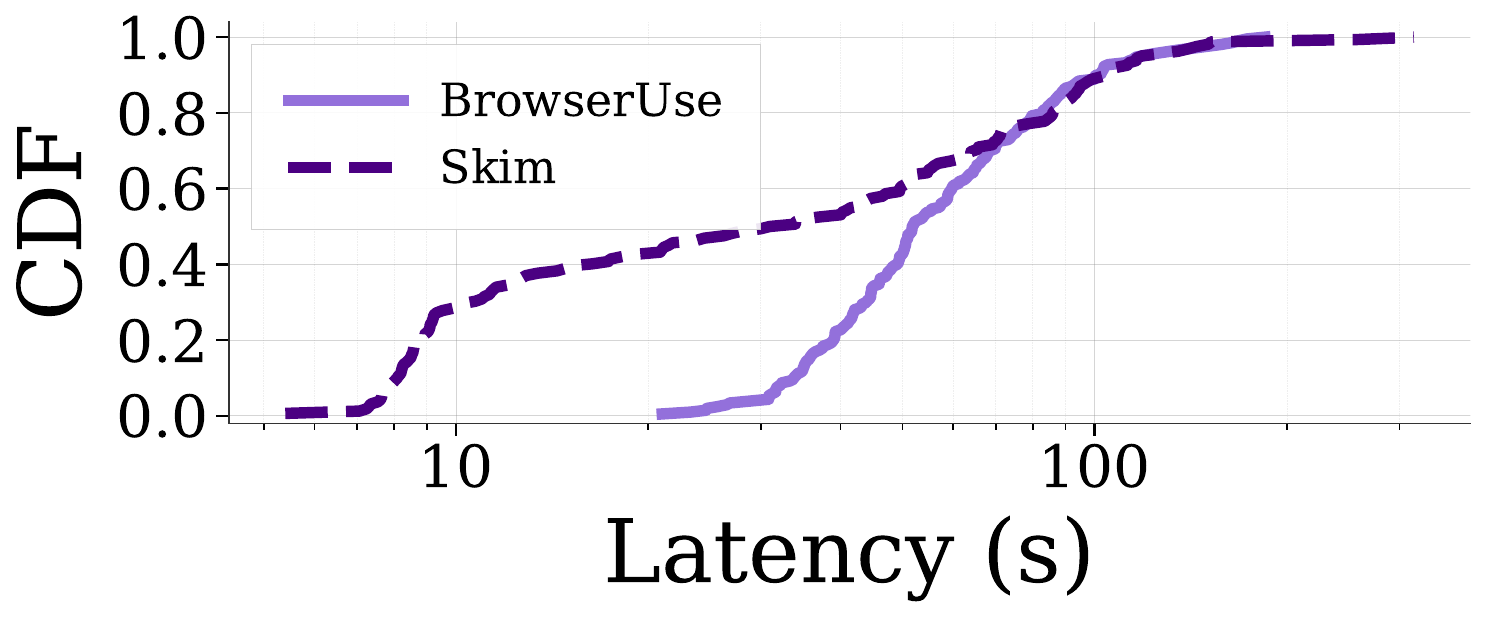}
    \end{minipage}

    \tightcaption{End-to-end task latencies for WebVoyager (left), AgentOccam (middle), and BrowserUse (right) across all tasks.}
    \label{fig:end-to-end-latencies}
\end{figure*}
\begin{figure*}[t]
    \centering

    \begin{minipage}{0.32\textwidth}
        \centering
        \includegraphics[width=\linewidth]{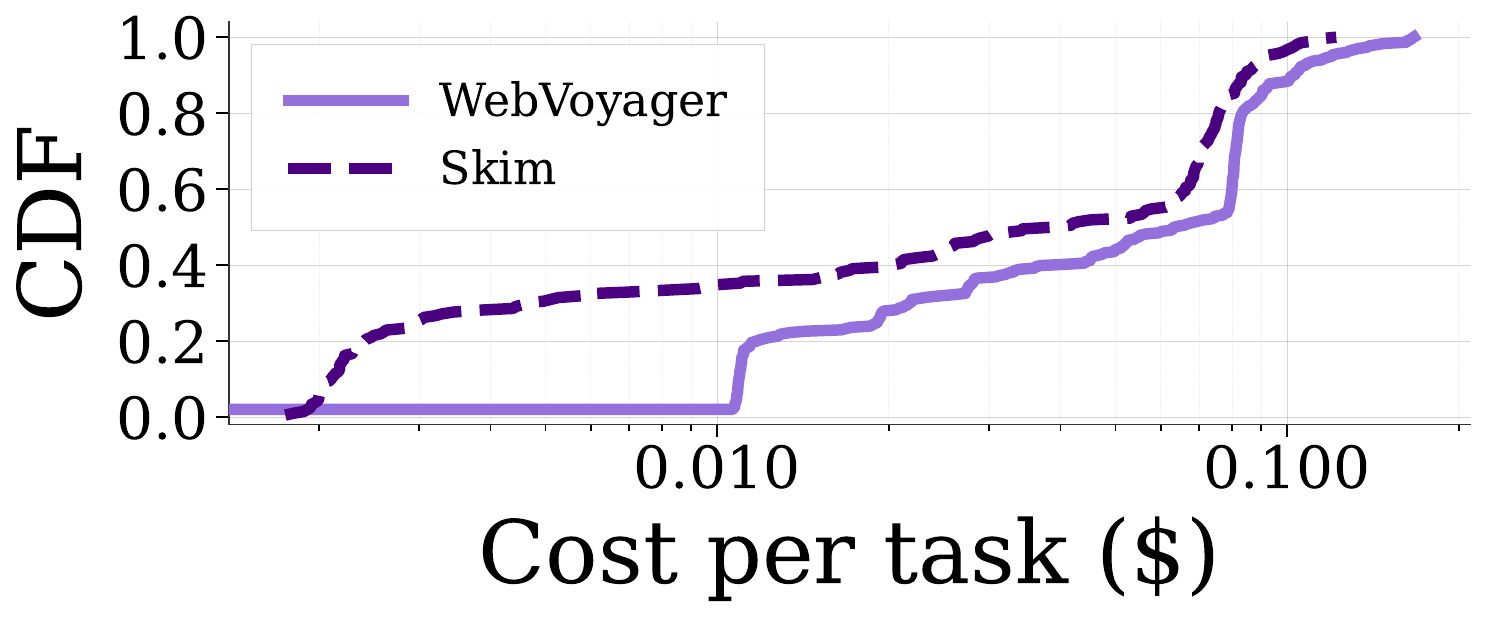}
    \end{minipage}
    \hfill
    \begin{minipage}{0.32\textwidth}
        \centering
        \includegraphics[width=\linewidth]{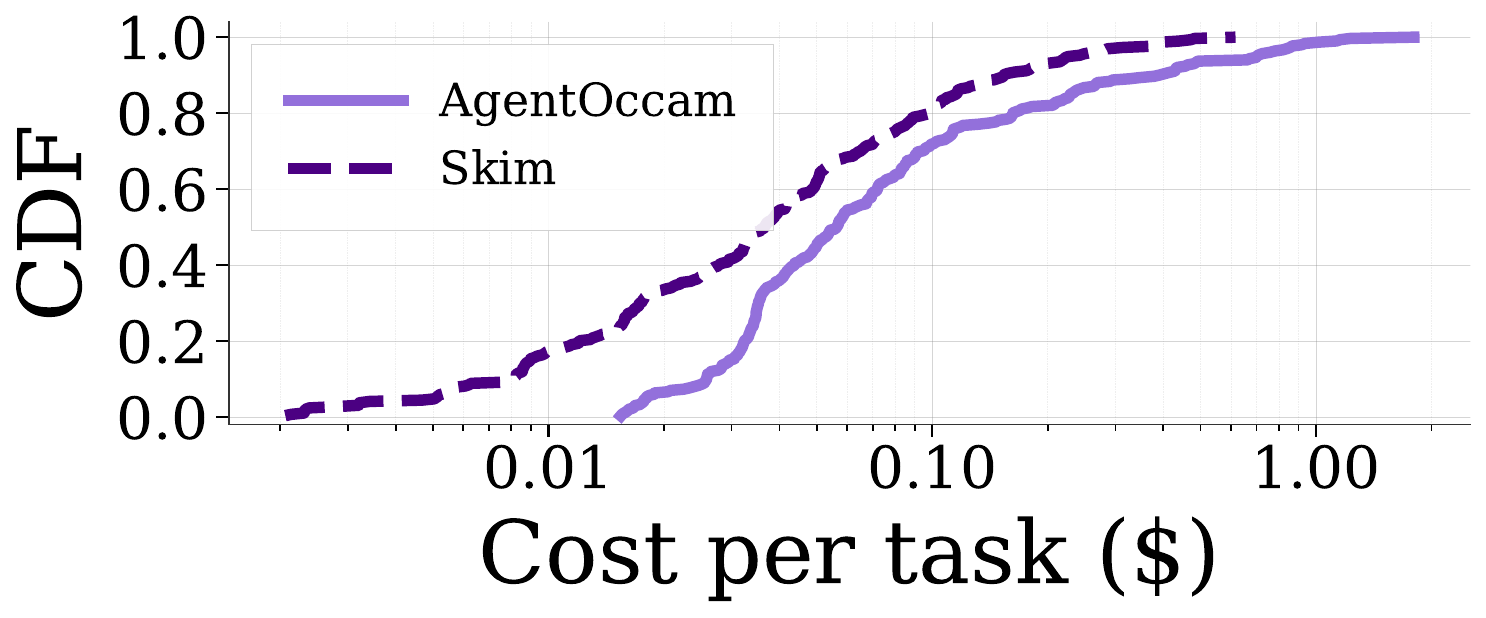}
    \end{minipage}
    \hfill
    \begin{minipage}{0.32\textwidth}
        \centering
        \includegraphics[width=\linewidth]{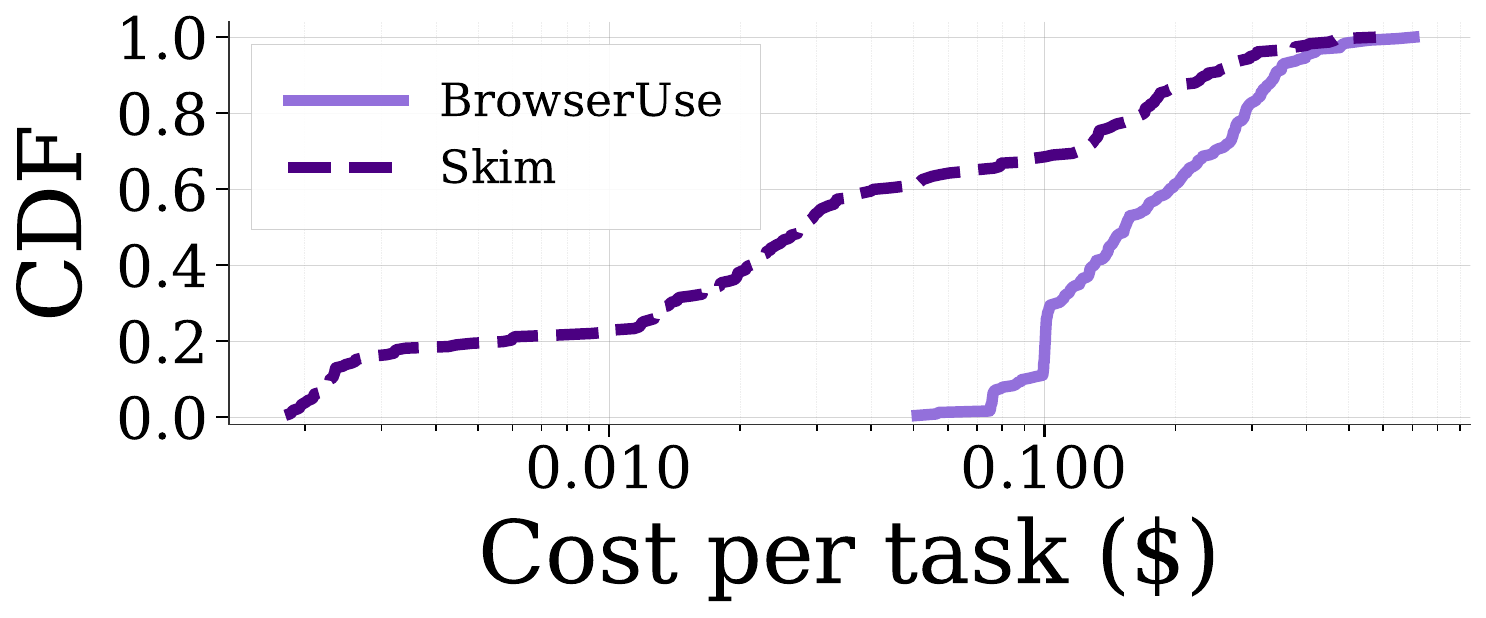}
    \end{minipage}

    \tightcaption{End-to-end task costs for WebVoyager (left), AgentOccam (middle), and BrowserUse (right) across all tasks.}
    \label{fig:end-to-end-costs}
\end{figure*}

\subsection{Methodology}
\begin{sloppypar}
We evaluate \sys atop three representative ReAct-based web agents: WebVoyager~\cite{webvoyager}, AgentOccam~\cite{agentoccam}, and BrowserUse~\cite{browseruse}. WebVoyager is a multimodal agent over rendered screenshots, serving as a strong baseline for general-purpose live-web navigation. AgentOccam emphasizes lightweight reasoning and action efficiency through aggressive state pruning over text-based DOM observations, representing cost-conscious agent designs. BrowserUse is a production-oriented browser-automation framework that tightly couples LLM reasoning with browser-control primitives, reflecting the deployment-focused agents used in retrieval and automation pipelines. Together, these systems span a range of reasoning styles, observation modalities, and execution overheads.
\end{sloppypar}

We consider two representative benchmarks. WebVoyager~\cite{webvoyager} spans 15 real-world websites with open-ended navigation, retrieval, and structured extraction tasks. WebShop~\cite{webshop} provides a shopping environment derived from Amazon product data, focused on multi-step product search and selection. Because evaluating live ReAct agents at scale is both computationally expensive and time-consuming (browsers and frontier models at each step), we conduct our experiments on a representative subset of 300+ tasks randomly drawn across the two benchmarks. Each task is executed on every agent backend with \sys. This scale captures a broad range of websites, task structures, and execution behaviors while keeping evaluation practically tractable. Ablations and detailed analyses are reported on AgentOccam and WebVoyager, though all trends hold for other combinations. Metrics include end-to-end latency, per-task dollar cost, and task success rate. Latency and cost are measured over complete task executions including failures and cascades.

\subsection{Main Results}

\begin{figure}[t]
    \centering
    \begin{minipage}{0.48\columnwidth}
        \centering
        \includegraphics[width=\linewidth]{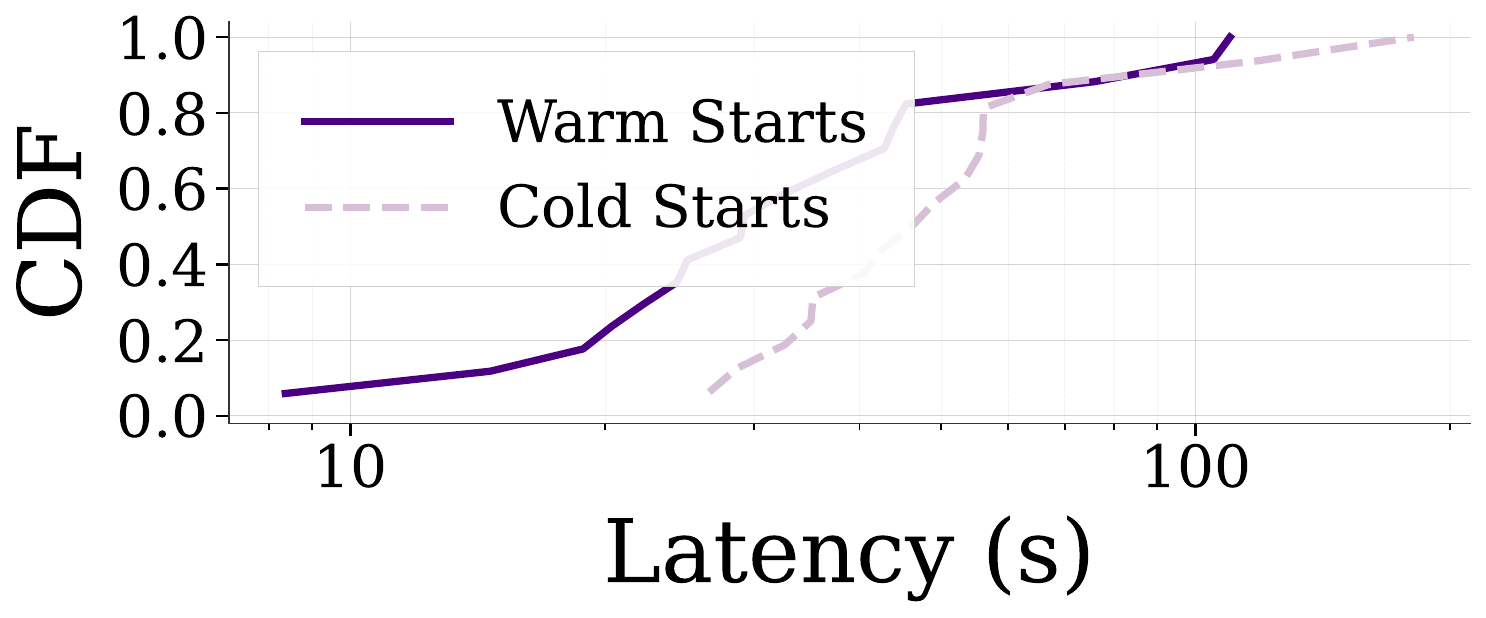}
        \tightcaption{Warm start savings for WebVoyager.}
        \label{fig:warm-start-webvoyager-savings}
    \end{minipage}
    \hfill
    \begin{minipage}{0.48\columnwidth}
        \centering
        \includegraphics[width=\linewidth]{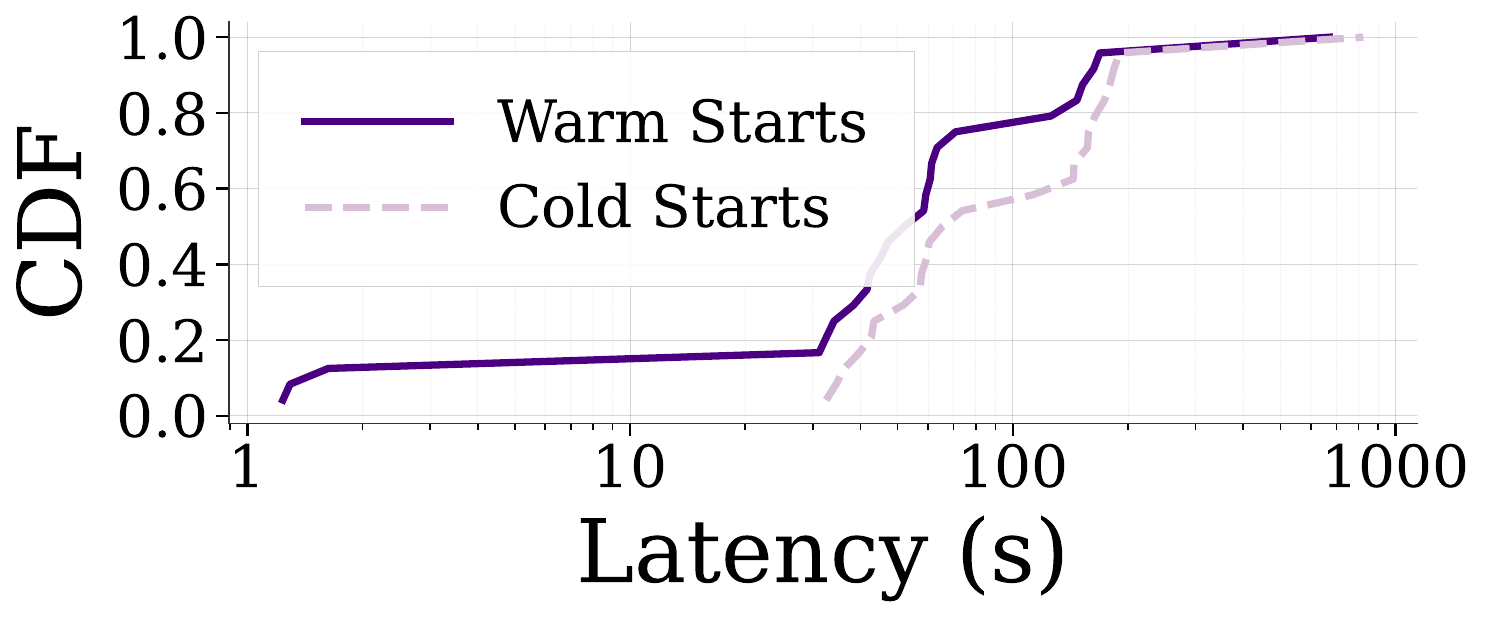}
        \tightcaption{Warm start savings for AgentOccam.}
        \label{fig:warm-start-agentoccam-savings}
    \end{minipage}
\end{figure}

Figures~\ref{fig:end-to-end-latencies}-\ref{fig:end-to-end-costs} report per-task latency and cost CDFs against each baseline. \sys achieves its largest gains on tasks that complete on the speculative fast path: navigation collapses to direct URL synthesis, repeated frontier-model inference is avoided, and extraction runs over condensed HTML with lightweight models. These tasks complete at a small fraction of full-ReAct latency and cost. Gains diminish in the tail as tasks trigger fallback to the underlying ReAct agent, either because verification rejected the fast path or because no template matched. However, \sys often preserves partial navigational progress through warm-start URLs, so even fallbacks frequently outperform cold-start ReAct (Figures~\ref{fig:warm-start-webvoyager-savings}-\ref{fig:warm-start-agentoccam-savings}). At the far tail, performance converges to the baseline in two cases: when \sys determines that no fast path applies and routes directly to the ReAct backend, or when the speculative attempt reaches a URL without useful progress; for the latter, overhead is bound by the failed attempt and verification stage.

Despite the fallback cases, \sys preserves end-to-end accuracy within noise of full ReAct (Table~\ref{t:accuracy-comparison}) while substantially reducing median latency and cost. Two properties drive this. First, speculative execution follows trajectories grounded in real structural patterns from offline profiling -- URL templates, search behavior, pagination structure, answer schemas -- that are tied to the site's core purpose and remain stable even as underlying content changes, rather than unconstrained lightweight reasoning over arbitrary pages. Second, every fast-path result is gated through verification before commitment, catching missing fields, unexpected page structures, failed extraction, and stale templates. When verification rejects, \sys reverts to the full ReAct agent rather than committing a wrong answer, so failures degrade gracefully into standard agent behavior.
\begin{table}[t]
    \centering
    \small
    \begin{tabular}{lcc}
        \toprule
        \textbf{Agent} & \textbf{Our accuracy} & \textbf{Agent accuracy} \\
        \midrule
        WebVoyager & 40.6\% & 37.6\% \\
        AgentOccam & 52.0\% & 49.6\% \\
        BrowserUse & 45.6\% & 45.0\% \\
        \bottomrule
    \end{tabular}
    \vspace{8pt}
    \tightcaption{Accuracies for \sys and default web agents.}
        \label{t:accuracy-comparison}
\end{table}

\para{Aggregate mode.} \sys{}'s cost savings can be repurposed to run multiple speculative trials per task within the baseline's single-execution budget, with results aggregated across trajectories. For instance, on the WebVoyager benchmark with AgentOccam, this enables \sys to run an average of four additional trajectories per task, with accuracy wins varying by sampling strategy: upper-bound wins (i.e., assuming oracle selection of the best trial) reach 16.7 percentage points, while majority vote yields 4.2-percentage-point improvements.

\begin{figure}[t]
    \centering
    \includegraphics[width=0.9\columnwidth]{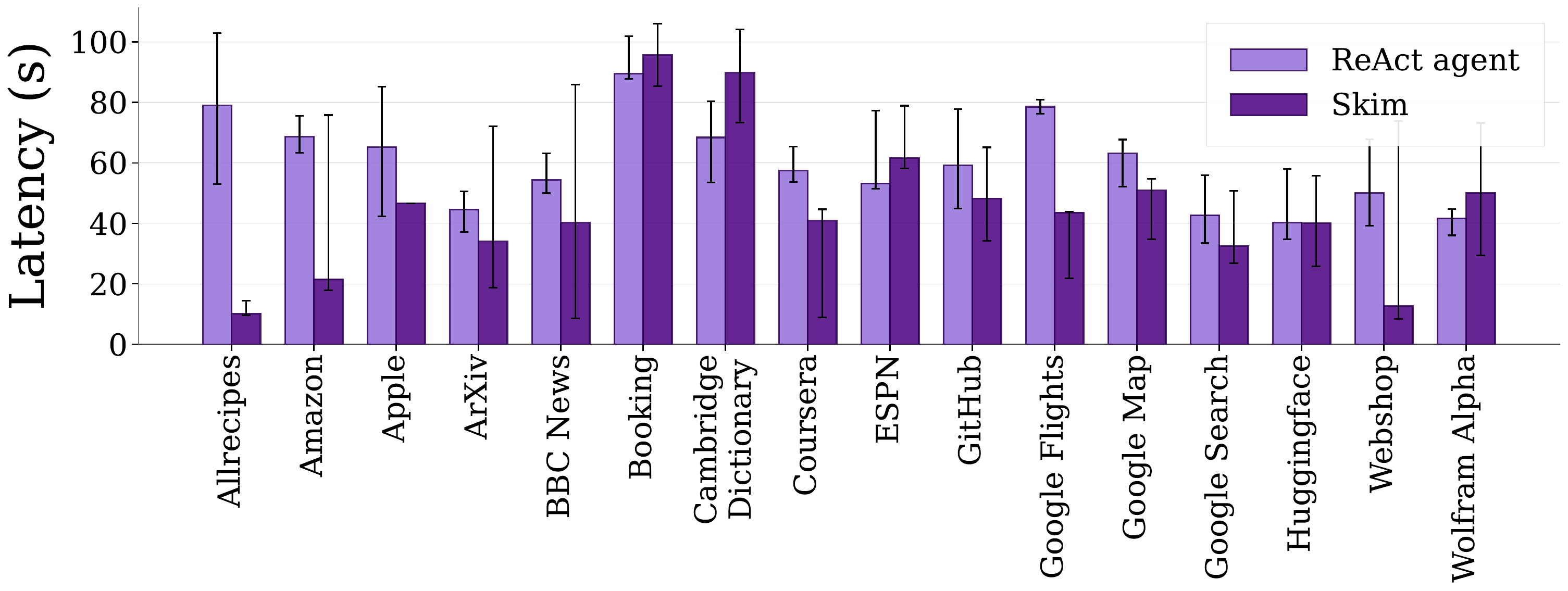}
    \tightcaption{Median end-to-end latency by website; error bars show 25th/75th percentiles. }
    \label{fig:eval-latency-breakdown-by-task}
\end{figure}

\para{Per-site speedups.}
Figure~\ref{fig:eval-latency-breakdown-by-task} reports median end-to-end latency across websites with heterogeneous retrieval and interaction structure. \sys{} achieves the largest speedups on sites whose workflows are dominated by structured search, filtering, and retrieval over stable page schemas. In these settings, speculative execution collapses multi-step browser interaction into direct URL synthesis followed by lightweight extraction over cleaned HTML, bypassing most of the iterative ReAct loop and avoiding many sequential browser actions and frontier-model inference calls. These gains are not tied to any single benchmark or domain; they emerge whenever a site exposes reusable navigation conventions and answer-bearing content resolvable through direct retrieval.

Latency improvements diminish on sites where speculation more frequently triggers fallback. These cases generally arise when site structure is difficult to parameterize through stable templates alone -- heavily JavaScript-driven interfaces, aggressive bot-detection, multimodal content requiring visual reasoning, or search semantics highly sensitive to exact query phrasing -- making direct retrieval less reliable and increasing verification failure and cascade. A small number of tasks also show speculative execution slightly increasing latency relative to the baseline, primarily when the warm start provides little useful navigational progress or when nondeterminism in LLM-based agent execution causes the baseline agent to converge unusually quickly. Even then, the overhead from failed speculation remains small relative to the cost of full browser-agent execution.

\para{Decomposing the savings gap.}
Two structural factors gap our savings from the hand-engineered ceiling. First, \sys pays a generalization tax of ~5-6s per task for routing, URL synthesis, and capability prediction. Second, only 12.6-45.3\% complete on the fast path; the rest cascade with warm-started ReAct, which captures most of the remaining gain through prefix-sharing rather than direct speculation success.
\begin{figure}[t]
    \centering
    \includegraphics[width=0.9\columnwidth]{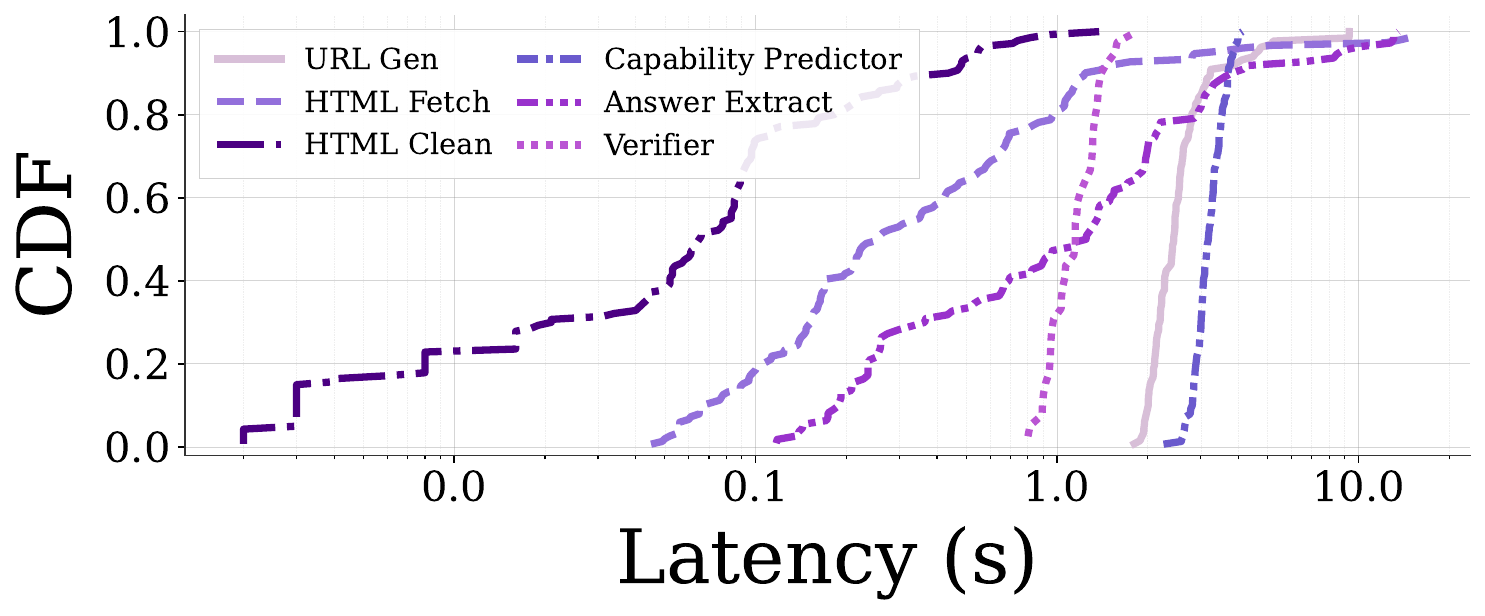}
    \tightcaption{Per-task latency breakdown}
    \label{fig:eval-latency-breakdown}
\end{figure}

\para{Impact of warm starts.}
Figures~\ref{fig:warm-start-webvoyager-savings} and
\ref{fig:warm-start-agentoccam-savings} compare fallback latency
under warm- vs cold-started ReAct execution. Warm starts reduce fallback latency by
seeding the downstream agent with the URL reached during
speculative execution instead of restarting from the homepage.
Even when the fast path fails, it often resolves substantial
parts of the trajectory (search, filtering, pagination, query
construction), so fallback resembles local recovery from a
near-correct state rather than exploratory browsing from
scratch. The latency distributions show that this substantially
compresses the tail of expensive fallback trajectories across
both agents. This reflects the prefix-sharing property:
incorrect speculative trajectories remain close enough to the
correct path that ReAct recovers with little additional
exploration, allowing \sys{} to preserve latency gains even on
cascade.

\subsection{Detailed Analysis}

\para{Latency breakdown.}
We decompose per-task latency across system components to identify where time is spent along the speculative execution path. Figure~\ref{fig:eval-latency-breakdown} reports stage-level wall-clock latency for fast-path-completing tasks across six stages: routing and capability prediction, URL synthesis, HTTP fetch, HTML cleaning, lightweight extraction, and verification. The dominant costs arise not from web access itself, but from the lightweight semantic stages responsible for interpreting and parameterizing tasks: routing and capability prediction take $\sim$3s per task, and URL synthesis another $\sim$2--3s for most queries (with a small long tail reaching $\sim$9s). By contrast, the systems-oriented stages are extremely cheap: HTTP fetches typically complete within 100--300ms, and HTML cleaning is usually below 100ms. This highlights a central property of read-only web tasks: once the correct destination URL is known, retrieving and condensing the relevant content is inexpensive relative to discovering that URL through exploratory browser interaction.

Extraction latency exhibits the largest variance: simple schema-constrained retrievals complete in well under a second, while harder synthesis-style tasks occasionally require 8--14s of local-model reasoning. Verification adds only $\sim$1s of overhead, preserving the design goal that correctness checking remain cheaper than the speculative execution it validates. For cascaded tasks, the full ReAct loop dominates tail latency, but failed fast paths frequently still preserve useful navigational progress through warm-start URLs, allowing the downstream agent to resume near the relevant destination page rather than restarting from the homepage. Per-task cost follows the same asymmetry: fast-path tasks are dominated by inexpensive local-model inference and HTTP execution, while cascaded tasks inherit the frontier-model cost profile of full browser-agent execution.

\begin{figure}[t]
    \centering
    \includegraphics[width=0.9\columnwidth]{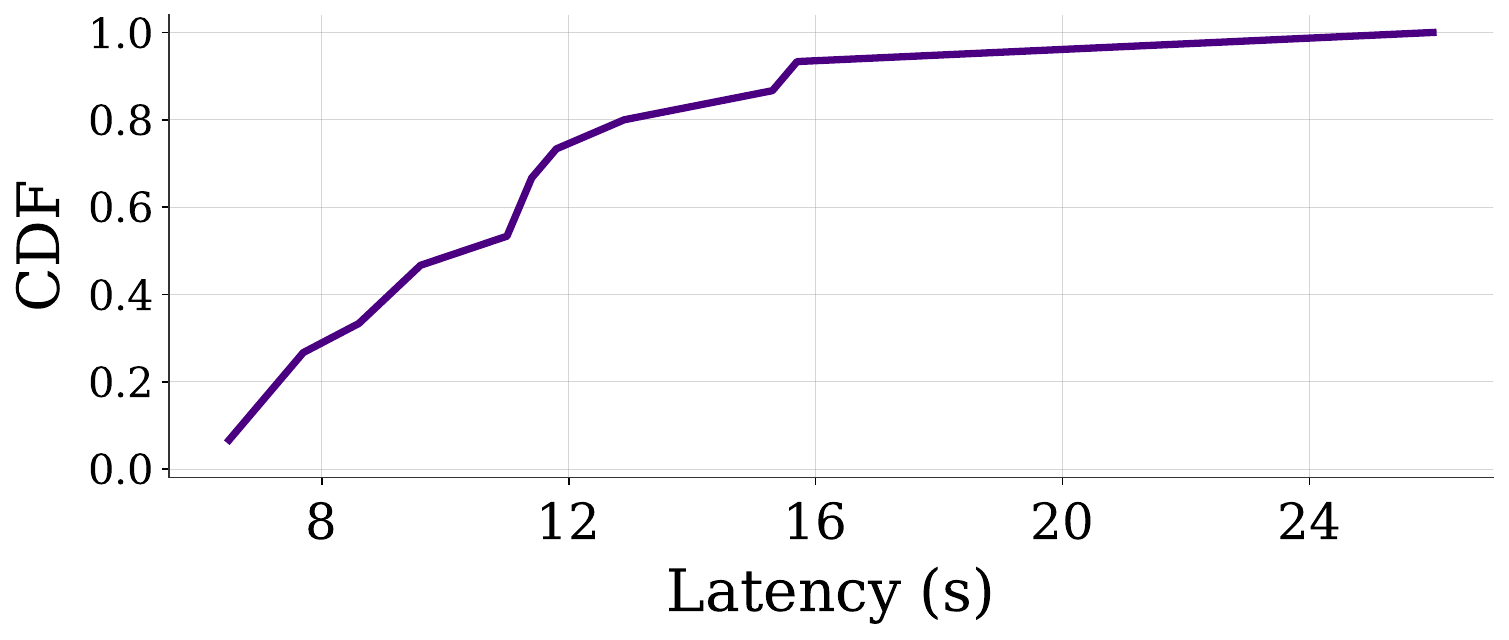}
    \tightcaption{Offline profiling latency per site. }
    \label{fig:offline-profiling-cdf}
\end{figure}

\para{Verification cost vs. quality.} The verifier must satisfy two goals: it must remain cheap enough that verification does not erase the fast path's savings, while being conservative enough that incorrect speculative answers rarely commit. We therefore explicitly bias the verifier toward rejection and aggressive cascade rather than risking false accepts: borderline or weakly-supported speculative outputs are escalated to heavier execution tiers instead of being committed directly.

We compare \sys's lightweight verifier against a frontier-model verifier that scores candidate answers against the full DOM. The lightweight verifier is 11.5$\times$ cheaper per call while achieving 82.0\% precision and 86.2\% recall (F1 = 0.84, accuracy = 86.9\%). Reliability is highest on sites with structured, schema-friendly outputs and lower on sites with heavier client-side rendering or looser semantics; importantly, these are also the sites where speculative execution more frequently triggers fallback, so verifier uncertainty naturally translates into cascade rather than silently committed errors.

\para{Offline profiling pipeline.}
Figure~\ref{fig:offline-profiling-cdf} shows the one-time cost
of offline profiling. Profiling completes within 10s for
${\sim}40$\% of sites and within 12s for ${\sim}60$\%, with a
small tail toward 16s on structurally complex sites requiring
browser escalation, dynamic rendering, or bot-detection
handling. The cost stays this low because the profiler captures
the small set of URL templates and answer schemas that cover
common-case tasks rather than enumerating every page or query
the site supports; tasks that fall outside profiled patterns
are detected at match time and routed to the ReAct fallback, so
coverage gaps cost nothing in correctness. The cost is amortized
across all future tasks on the site and does not contribute to
online query latency. Profiles are continuously validated
through schema and content-consistency checks at runtime; stale
profiles are regenerated when structural drift is detected.

\section{Related Work}
\label{sec:related}

To our knowledge, \sys is the first system to treat website
structure as a speculative execution primitive for accelerating
web agents. Prior work instead improves agent
policies and benchmarks, accelerates inference and execution, or
exploits webpage structure for extraction and compression.

\para{Web agents and benchmarks.}
Recent work has rapidly expanded the capabilities of LLM-based
web agents and the realism of their evaluation environments.
Early systems such as WebGPT~\cite{webgpt} and
ReAct~\cite{react} established the paradigm of interleaving
reasoning with actions over external environments, with later
work improving planning, grounding, prompting, observation
representations, skill reuse, program synthesis, and benchmark
realism~\cite{webagent,mind2web,webshop,webarena,weblinx,webvoyager,gaia,odysseys,webxskill,seeact,agentoccam,molmoweb,browseruse},
alongside interface grounding and perception
work~\cite{showui,uground,pixels2uiactions}. The closest
antecedent is WebAgent~\cite{webagent}, which synthesizes
executable programs for websites.
Its motivation is adjacent to ours, but the core question
differs: WebAgent improves the policy operating inside the online
loop, whereas \sys asks when that loop can be bypassed entirely
through verified site-structured fast paths. More broadly, prior
systems strengthen the underlying agent but still execute a
mostly uniform browser-and-LLM loop online; \sys instead targets
the execution substrate itself, speculating away browser
interaction and frontier-model use when site structure makes a
cheaper path safe.

\para{Speculation and efficient inference.}
Another line of work targets efficiency through model cascades,
routing, and speculative execution. FrugalGPT~\cite{frugalgpt}
learns cascades over models to reduce cost, while speculative
decoding accelerates generation by drafting with a smaller model
and verifying with a larger one~\cite{speculative_decoding}.
Related approaches reduce latency through auxiliary drafting,
branching, caching, or verification
work~\cite{contextbudget,readmorethinkmore,dsp,speculative_actions,speccache}.
These methods operate primarily at the level of tokens, context,
or next-action prediction within an otherwise unchanged agent
loop. \sys instead speculates over \emph{site-structured
execution paths}: direct URLs, HTTP-only retrieval, and
schema-constrained extraction. Rather than adding speculative
computation ahead of the same browser trajectory, \sys removes
browser interaction and entire ReAct steps whenever verification
accepts the fast path.

\para{Wrapper induction and web extraction.}
\sys also relates to classical wrapper induction for
semi-structured webpages~\cite{kushmerick2000wrapper} and modern
work on HTML understanding, DOM reasoning, pruning, and
region-level abstractions~\cite{html_understanding,domqnet,prune4web,region4web,contractskill}.
These systems exploit recurring webpage structure for extraction
or observation compression \emph{after} the agent reaches the page.
\sys operates one level earlier: it uses site structure not only
for extraction, but also for URL synthesis, retrieval-mode
selection, speculative execution, and runtime verification. In
this sense, wrapper-like structure becomes a verified fast path
rather than the sole execution mode.
\section{Conclusion}


We present \sys, a speculative execution framework that makes web agents substantially faster and cheaper without sacrificing the coverage of full ReAct execution. \sys exploits the structural regularity of purpose-built websites to replace much of the default browser-and-frontier-model loop with direct retrieval, lightweight extraction, and cheap verification, while preserving robustness through warm-started fallback when speculation fails. In accelerate mode, \sys reduces median per-task cost by 1.9$\times$ and latency by 33.4\% across three diverse ReAct backends while preserving end-to-end accuracy. In aggregate mode, the savings support an average of four additional speculative trials per task, improving accuracy by up to 16.7 percentage points (4.2 pp with majority vote). 

\bibliographystyle{plain}
\bibliography{references}


\end{document}